`
\documentclass[12pt]{iopart}
\usepackage{iopams}
\usepackage[T1]{fontenc}
\usepackage[utf8]{inputenc}
\usepackage{amssymb}
\usepackage[numbers]{natbib}
\usepackage{algorithm}
\usepackage{algpseudocode}
\usepackage[skip=0pt]{subcaption}
\usepackage{tabularx}
\usepackage{multirow}
\usepackage[dvipsnames]{xcolor}
\usepackage{homemade_commands}

\makeatletter
\newcommand{\printfnsymbol}[1]{%
  \textsuperscript{\@fnsymbol{#1}}%
}
\makeatother

\usepackage{hyperref}
\usepackage[normalem]{ulem}

\definecolor{lightblue}{RGB}{82,82,183}
\hypersetup{
    colorlinks=true,
    citecolor=lightblue,
    linkcolor=lightblue,
    urlcolor=lightblue
}

\newcommand{\Cedric}[1]{\textcolor{black}{#1}}
\newcommand{\CedricTwo}[1]{\textcolor{black}{#1}}

\newcommand{\phimax}{\varphi_{\max}}
\newcommand{\pdrop}{p_{\textrm{drop}}}
\newcommand{\pshuffle}{p_{\textrm{shuffle}}}
\newcommand{\paug}{p_{\textrm{aug}}}
\newcommand{\trot}{\theta_{\textrm{rot}}}

\begin{document}

\title[Data augmentation for learning predictive models on EEG]{Data augmentation for learning predictive models on EEG: a systematic comparison}

\author{Cédric Rommel, Joseph Paillard, Thomas Moreau \& Alexandre Gramfort}

\address{Université Paris-Saclay, Inria, CEA, Palaiseau, 91120, France}
\ead{\{firstname.lastname\}@inria.fr}
\vspace{10pt}

\begin{abstract}
%
\textit{\CedricTwo{Objective:}} The use of deep learning for electroencephalography (EEG) classification tasks has been rapidly growing in the last years, yet its application has been limited by the relatively small size of EEG datasets.
Data augmentation, which consists in artificially increasing the size of the dataset during training,
\Cedric{can be employed to alleviate this problem.}
While a few augmentation transformations for EEG data have been proposed in the literature, their positive impact on performance \Cedric{is often evaluated on a single dataset and compared to one or two competing augmentation methods.}
\Cedric{This work proposes to better validate the existing data augmentation approaches through}
a unified and exhaustive analysis.
%
%
\textit{\CedricTwo{Approach:}}
\Cedric{We compare quantitatively}
\Cedric{13 different augmentations with two different predictive tasks, datasets and models, using three different types of experiments.}
%
%
\textit{\CedricTwo{Main results:}}
\Cedric{We demonstrate that employing the adequate data augmentations can bring up to 45\% accuracy improvements in low data regimes compared to the same model trained without any augmentation.}
\CedricTwo{Our experiments also show that there is no single best augmentation strategy, as the good augmentations differ on each task.}
\textit{\CedricTwo{Significance:}}
Our results highlight the best data augmentations to consider for sleep stage classification and motor imagery brain-computer interfaces.
\CedricTwo{More broadly, it demonstrates that EEG classification tasks benefit from adequate data augmentation.}
\end{abstract}

%
\vspace{2pc}
\submitto{\JNE}
%
%
\ioptwocol
%

\section{Introduction}
%
%
Decoding the brain electrical activity is a great scientific challenge for both clinicians and researchers seeking a better understanding of brain dynamics.
Recent attempts to leverage deep learning for this difficult task have shown promising results \cite{roy_deep_2019}.
These new methods have led to performance gains in a wide range of clinically relevant tasks, such as the automatic sleep stage classification from polysomnographic recordings \cite{chen_deep_2020,xsleepnet,perslev2021u,chambon_deep_2017}.
The ambition of deep learning models is to automatically learn relevant representations from high dimensional data such as EEG \cite{ang_filter_2012,gemein_machine-learning-based_2020}, while previously used brain decoding methods relied on prior knowledge and handcrafted features \cite{ramoser_optimal_2000}.
Doing so, deep learning approaches require a less sharp understanding of the underlying neurophysiology and are thus more versatile. Yet, this comes at the cost of
\Cedric{requiring large training datasets.}

%

Indeed, most of the breakthroughs in deep learning have been enabled by large datasets such as \textit{ImageNet} \cite{russakovsky_imagenet_2015}. Unfortunately, similar datasets do not exist in neuroscience as labeled brain data remains comparatively scarce.
Labelling EEG recordings requires a high expertise, is time consuming and can sometimes be inaccurate due to the bias introduced by the human annotator \cite{rosenberg_american_2013}.
A second obstacle to the application of deep learning in neuroscience is the high inter-subject variability that is inherent in brain signals \cite{clerc_brain-computer_2016}.
Along with the lack of data, this property makes the generalization on unseen subjects particularly difficult.
Without proper regularization, both of these problems can hinder generalization performance and lead to overfitting.

%

To mitigate the small scale of the neuroscience databases, a promising direction is the use of data augmentation \cite{krizhevsky_imagenet_2012, specaugment_park_2019, nlpaug}. Data augmentation allows to increase artificially the size of the training set by adding new synthetic examples.
These examples are generated by randomly transforming existing ones in a label-preserving way.
Doing so, data augmentation helps the decision function to become invariant to the transformation enforced, thus softly reducing the hypothesis space of the training problem.
Consequently, it can be interpreted as a regularization method that induces a useful bias by preventing the model from focusing on irrelevant features~\cite{chen_group-theoretic_2020}, which in the end makes it less prone to overfitting~\cite{srivastava_dropout_2014}.
%

\Cedric{Although data augmentation is a well-established method in computer vision, and despite a number of recent studies, data augmentation is still under-explored for EEG data. Among the recent studies listed in \autoref{tab:data_augs}, some propose to perturb the data in either the spatial domain of sensors \cite{krell_rotational_2017, deiss_hamlet_2018, saeed_learning_2020}, the frequency domain of signals \cite{schwabedal_addressing_2019, cheng_subject-aware_2020} or the time domain \cite{wang_data_2018, mohsenvand_contrastive_2020, rommel2021cadda}.}
\Cedric{While some reviews cover data augmentation for EEG \citep{roy_deep_2019,lashgari2020data, he2021data}, they mainly focus on summarizing results from the literature without carrying out extensive new experiments.}

\paragraph{Contributions}
\Cedric{In this paper, we propose to better validate the main existing EEG data augmentation methods listed in \autoref{tab:data_augs} through a unified and exhaustive analysis in the context of sleep stage classification and motor imagery brain-computer interfaces (BCI) \cite{clerc_brain-computer_2016}.}
\Cedric{In total, we compare 13 different augmentations with two different predictive tasks, datasets and models.}
\Cedric{The objectives of our experiments are three-fold:
(i) to evaluate the impact of the magnitude of each transformation, (ii) to compare the benefit of each augmentation with different training set sizes, and (iii) to highlight how the effects of augmentations vary across data classes.}
\Cedric{We organize our analysis as follows.}
\Cedric{First, we outline in \autoref{sec:protocol} the experimental setting and protocol used to tune and compare all data augmentations.}
\Cedric{Then, we describe in more details the rationale of each augmentation and present the experimental results for time domain augmentations (\autoref{sec:time-tfs}), frequency domain augmentations (\autoref{sec:frequency-tfs}) and spatial domain augmentations (\autoref{sec:spatial-tfs}).}
\Cedric{Finally, we summarize our findings and draw more general conclusions in \autoref{sec:gen-discussion}.
The code used in our experiments has been available in an open-source repository\footnote{\url{https://github.com/eeg-augmentation-benchmark/eeg-augmentation-benchmark-2022}}.}



\begin{table*}[t]
    \centering
    \Cedric{
    \begin{tabularx}{\textwidth}{p{30mm} c p{40mm} X}
        \hline
        Augmentation & Type & Reference & Short description \\ \hline\hline
        \texttt{FTSurrogate} & F & \citet{schwabedal_addressing_2019} & Randomize Fourier phases of all channels. \\ \hline
        \texttt{BandstopFilter} & F & \citet{mohsenvand_contrastive_2020}, \citet{cheng_subject-aware_2020} & Randomly filter a small frequency band of all channels. \\ \hline
        \texttt{FrequencyShift} & F & \citet{rommel2021cadda} & Randomly translate all channels PSD by small shift. \\ \hline
        \texttt{GaussianNoise} & T & \citet{wang_data_2018} & Add Gaussian white noise to the signals. \\ \hline
        \texttt{SmoothTimeMask} & T & \citet{mohsenvand_contrastive_2020} & Randomly pick a portion of the signal and set it to zero. \\ \hline
        \texttt{SignFlip} & T & \citet{rommel2021cadda} & Randomly flip the sign of all channels. \\ \hline
        \texttt{TimeReverse} & T & \citet{rommel2021cadda} & Randomly reverse the axis of time in all channels. \\ \hline
        \texttt{ChannelsSymmetry} & S & \citet{deiss_hamlet_2018} & Randomly swap signals from right hemisphere to left hemisphere and \textit{vice-versa}.\\ \hline
        \texttt{ChannelsDropout} & S & \citet{saeed_learning_2020} & Randomly pick a given number of channels and set their signals to zero. \\ \hline
        \texttt{ChannelsShuffle} & S & \citet{saeed_learning_2020} & Randomly pick a given number of channels and permute their signals. \\ \hline
        \texttt{SensorsRotation} & S & \citet{krell_rotational_2017} & Interpolate channels signals on randomly rotated positions. \\ \hline
    \end{tabularx}
    \caption{Data augmentation methods studied in this work. Types stand for Frequency (F), Time (T) and Spatial (S) transformations.}
    \label{tab:data_augs}
    }
\end{table*}

\section{Experimental protocol}
\label{sec:protocol}

The experiments are conducted on two different tasks: sleep stage classification and motor imagery classification in the context of BCI. Both tasks and datasets are described in \autoref{subsec:EEG_tasks}.
\autoref{subsec:experiments} presents the common experimental protocols used to study each data augmentation
\Cedric{and \autoref{sec:implem-details} provides implementation details for reproducibility purposes.}


\subsection{EEG classification tasks}
\label{subsec:EEG_tasks}

\subsubsection{Sleep stage classification}
\paragraph{Dataset and preprocessing} 
Sleep stage classification is essential to diagnose sleep disorders such as sleep apnea or insomnia.
It consists in classifying EEG windows of 30 seconds into 5 stages defined by the \emph{American Academy of Sleep Medecine} (\textit{AASM}) manual \cite{berry_aasm_2015}: Wake (W), Rapid Eye Mouvement (REM) and Non REM stages 1, 2 and 3 (respectively N1, N2 and N3).
This task is usually performed by sleep experts, which can be very time consuming and subjective, motivating the use of automatic classification systems when possible.
To evaluate the impact of data augmentation on automatic sleep stage classification, we used the \textit{SleepPhysionet} dataset \cite{goldberger_physiobank_2000}, which contains whole night polysomnographic recordings from 78 healthy subjects using two EEG channels: Fpz-Cz and Pz-Oz.
In this dataset, each signal's window has been annotated by trained technicians according to the \textit{Rechtschaffen and Kales} manual \cite{rechtschaffen_manual_1973}, which are re-asigned to the more recent stages from the \textit{AASM} manual.
As suggested in \cite{chambon_deep_2017}, a minimal data preprocessing is applied, consisting in a low-pass filter with a cutoff frequency of $30$\,Hz, followed by a standardization step (each channel's signal is centered and scaled to have unit variance).

%
%
\paragraph{Splitting strategy}
EEG recordings have a strong inter-subject variability \cite{clerc_brain-computer_2016}.
Subsequently, to test the trained models in real conditions, some subjects must be set aside and used only for testing in order to avoid subject related information leakage \cite{roy_deep_2019}. 
To take this into account, the following splitting strategy has been defined.
First, the dataset is separated into $k$-folds, each of them containing different subjects. One fold is left out for testing and among the remaining $(k-1),\; 20\%$ of the subjects are used for validation. Finally, a subset of the remaining dataset is extracted using a stratified split and used for training.  
This last step allows to train the model in low data regime while maintaining the distribution of classes comparable to the validation and test sets. 

%
%
\paragraph{Model and training details}
Experiments are performed using a deep convolutional neural network that has been designed for sleep stage classification tasks \cite{chambon_deep_2017}.
It is trained using the Adam optimizer \cite{kingma_adam_2015} with a learning rate of $10^{-3}$.
The weighed cross-entropy loss is used to take into account the class-imbalance of the dataset. The batch size is set to 16 to preserve the stochasticity of the gradient descent in very low data regimes.  We train the model for 300 epochs using early-stopping on the validation loss with a patience of 30 epochs \cite{chambon_deep_2017}.

\subsubsection{BCI}
%
%
\paragraph{Dataset and preprocessing}
Likewise, experiments are carried out with the \textit{BCI IV 2a} dataset~\cite{brunner2008bci}.
It consists of recordings from 9 subjects using 22 EEG electrodes.
The subjects were asked to perform four motor imagery tasks, namely to imagine the movement of the left hand (class 1), right hand (class 2), both feet (class 3), and tongue (class 4).
This dataset was preprocessed using a bandpass filter between $4$\,Hz and $38$\,Hz followed by an exponential moving standardization as in \cite{schirrmeister_deep_2017}.
Then trials of 4.5 seconds are used as inputs.
Each trial starts 0.5 seconds before the cue that tells the subject to perform the motor imagery task and ends when the cue disappears.
%
%

\paragraph{Splitting strategy} According to the rules of the BCI competition \cite{brunner2008bci}, for each subject, our model is trained on the first session (or a fraction of it) and evaluated on the second session.
The experiment is repeated across all nine subjects.
%
%

\paragraph{Model and training details} Experiments are performed using a generic deep convolutional network \cite{schirrmeister_deep_2017}, as implemented in the library \textsc{Braindecode}~\cite{schirrmeister_deep_2017}.
This architecture is inspired by the success of Common Spatial Patterns (CSP) methods \cite{ramoser_optimal_2000}.
It can take advantage of the spatio-temporal structure of the data using spatial filters and convolutions across time.
Following the work of \cite{schirrmeister_deep_2017}, the network's training uses the AdamW optimizer  \cite{loshchilov_decoupled_2019} with a learning rate of $6.25 \times 10^{-4}$, batch size $64$, maximum of $1600$ epochs and early stopping on the validation error (patience of $160$ epochs).

\subsection{Experiments}
\label{subsec:experiments}

In this section, we describe the three types of experiments carried out with each EEG augmentation considered.

\subsubsection{Parameters selection}
\label{subsubsec:param_selection}
Several augmentations have a parameter that can be adjusted to control
how strongly the inputs are transformed.
For example, the Gaussian noise augmentation has a parameter $\sigma$ corresponding to the standard deviation of the distribution from which the noise is sampled.  The choice of the value of such a parameter is as important as the choice of the augmentation itself, as later depicted in our results (\Autoref{sec:frequency-tfs, sec:time-tfs, sec:spatial-tfs}).

This experiment unfolds in two steps: 1) narrowing down the range of parameter values and 2) carrying out a grid-search.
For the first step, an upstream manual exploration allows to estimate an upper bound above which the augmentation distorts too much the relevant information contained in the signal.
For instance, in the case of Gaussian noise, with $\sigma$ values greater than 0.2,
EEG signals become so noisy that the augmentation is systematically detrimental to the learning. 

For the second step, a grid-search is carried out using $11$ linearly spaced values within the aforementioned interval.
Since data augmentation is all the more efficient in low data regimes (as shown in our experiments from \Autoref{sec:freq-learning-curves, sec:time-learning-curve, sec:spatial-learning-curve}), we carried our parameter selection using a small balanced fraction of the initial datasets  (e.g. $2^{-7}$ for \textit{SleepPhysionet} dataset) to make potential improvements more apparent.
For each value in the grid, the accuracy metric is computed using a $10$-fold cross-validation.

\subsubsection{Learning curves}
\label{subsubsec:lr_curves}
The second experiment aims at comparing the benefits brought by different augmentation methods. 
To this end, for each augmentation operation, we compute a learning curve which shows the model's performance when it is trained on increasing fractions of the training set. 
Note that the same validation and test set are used for model selection and evaluation for all training set fractions, to ensure the learning curve points are comparable.
The results obtained with each data augmentation are then compared to a baseline, consisting in the same model trained with no data augmentation.

\subsubsection{Per class analysis}
\label{subsubsec:per_class}
Finally, to get a deeper understanding of the effects of data augmentations, we take a closer look at single points from the learning curve and analyze how effects vary depending on the classes. 
This observation is guided by the intuition that the invariances encoded by data augmentations might be more relevant for some classes than others. 
For example, the channel symmetry augmentation, which switches EEG channels from left and right hemispheres, is much more relevant for non-lateralized brain activities such as imagining tongue movements, whereas it is likely detrimental for lateralized functions, such as right- or left-hand movements.

The first experiments on the \textit{SleepPhysionet} dataset reveal that augmentations are systematically more helpful in low data regimes (\lcf \autoref{sec:freq-learning-curves}), and thus have a greater effect on underrepresented classes.
Since we are seeking to assess the effects of transformations on learned representations for each class, these experiments require class proportions to be equalized before training.
To do this, a subsampling step is added to the pre-processing pipeline, allowing to work with balanced data.
We choose to report results here in terms of 10-fold cross-validated F1-scores, which is a natural metric choice in class-wise analysis.

\subsection{Implementation}\label{sec:implem-details}

\subsubsection{Data augmentation}

Data augmentation consists in randomly applying an output-preserving transform to training samples.
More formally, for each input-output pair $(x, y)$ sampled from the training set, there is a probability $\paug$ of transforming the input using some transformation $T$ chosen \emph{a priori}.
$T$ should in theory be such that $P(y|T(x))$ is similar to $P(y|x)$.
We then feed the transformed input $T(x)$ to the model and train it to predict the original output $y$.
This procedure is carried on-the-fly, for each example of every mini-batch.
In all our experiments, we use a probability $\paug=0.5$ of augmenting each input and $1-\paug = 0.5$ of leaving them unchanged.

\subsubsection{Reproducible code}

The implementation of all data augmentations has been added to the open-source package \textsc{Braindecode}~\cite{schirrmeister_deep_2017}.
The data are automatically fetched thanks to MOABB~\cite{jayaram2018moabb} for the \textit{BCI IV 2a} dataset and MNE~\cite{GramfortEtAl2013a} for sleep physionet.
\Cedric{Moreover, the code used in all our experiments has been made available in an open-source repository\footnote{\url{https://github.com/eeg-augmentation-benchmark/eeg-augmentation-benchmark-2022}}.}

\section{Time domain augmentations}
\label{sec:time-tfs}
\subsection{\Cedric{Rationale of time domain augmentations}}
%
\begin{mdframed}[style=mystyle, frametitle=Gaussian noise]
    \Cedric{The \texttt{GaussianNoise} augmentation, proposed for example in \cite{wang_data_2018}, consists in adding Gaussian white noise to the recorded EEG signals.}
\end{mdframed}

\Cedric{In practice, a perturbation ${E(t) \sim \mathcal{N}(0, \sigma^2)}$ is sampled independently for each channel and acquisition time, and is added to the original signal $X$:
\begin{equation*}
    \mathtt{GaussianNoise}[X](t) = X(t) + E(t) \enspace.
\end{equation*}
Here $\sigma$ denotes the standard-deviation of the noise distribution.}
\Cedric{This parameter is interpreted as this transformation's magnitude, as the larger it is the more the original signal is distorted.}

\begin{figure*}[ht]
    \centering
    \includegraphics[width=\textwidth]{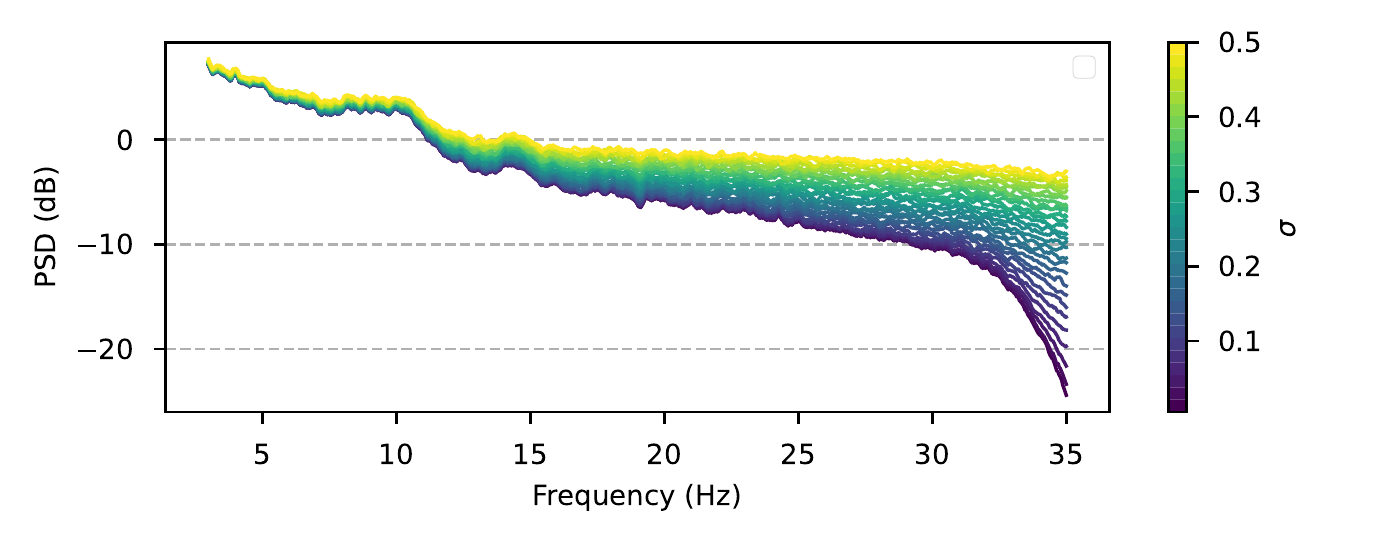}
    \caption{Effects of the addition of \texttt{GaussianNoise} on the PSD.
    Power spectra were averaged over N1 windows for one night of sleep from the \emph{SleepPhysionet} dataset.
    In polysomnograms, the power globally decreases as the frequency increases.
    Consequently, the \texttt{GaussianNoise}, which adds a constant amount of power across all frequencies, has a greater relative impact on higher frequencies.
    As we increase $\sigma$, a greater portion of the signal is hidden by the added noise.}
    \label{fig:gaussian-noise-illustration}
\end{figure*}

\Cedric{The motivation behind this augmentation is make the model more robust to noise in EEG recordings, as they are known to suffer from limited signal-to-noise ratio (SNR).}
%
%
As EEG signal power decreases with the frequency, this augmentation mostly preserves power band ratios at lower frequencies, which are instrumental in many EEG decoding tasks, such as sleep stage classification~\cite{Lajnef2015,chambon_deep_2017}.
\Cedric{On the contrary,}
by adding the same amount of power to all frequencies, the addition of white noise hides the information contained in high frequency bands,
as depicted in \autoref{fig:gaussian-noise-illustration}.
From this perspective, the effect of this transformation
is somehow analogous to a low-pass filter where the parameter $\sigma$
\Cedric{plays the role of}
a cut-off frequency.

\begin{mdframed}[style=mystyle, frametitle=Smooth time mask]
    \Cedric{The \texttt{SmoothTimeMask} augmentation, proposed in ~\cite{mohsenvand_contrastive_2020}, consists in replacing by zeros a portion of length $\Delta t$ of all channels, starting from a randomly sampled instant $t_{\textrm{cut}}$.}
\end{mdframed}

\begin{figure*}[ht]
    \centering
    \includegraphics[width=0.9\textwidth]{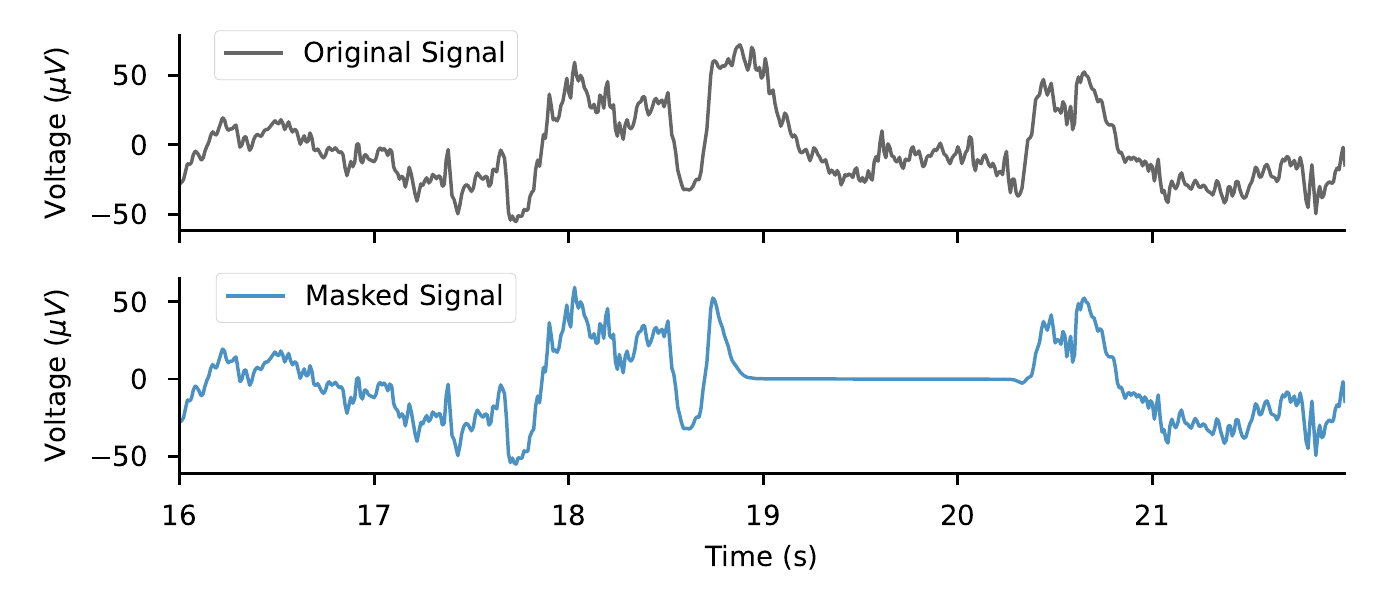}
    \caption{Effect of \texttt{SmoothTimeMask} on a time window from the \emph{SleepPhysionet} dataset. The mask length is $1.6sec$ and the transition between unchanged parts and the masked portion is smooth.}
    \label{fig:TimeMask}
\end{figure*}

\Cedric{In practice, the computation is carried out by multiplying the signal $X$ by a mask $m_\lambda$ made out of two opposing sigmoid functions of temperature $\lambda$:
\begin{eqnarray*}
    \mathtt{SmoothTimeMask}[X](t) := X(t) \cdot m_\lambda(t),\\
    m_\lambda(t) := \sigma_\lambda(t - t_{\textrm{cut}}) + \sigma_\lambda(t_{\textrm{cut}} + \Delta t - t)\\
    \sigma_\lambda(t) := \frac{1}{1+\exp{(-\lambda t)}} \enspace ,\\
    t_{\textrm{cut}} \sim \mathcal{U}[t_{\min}, t_{\max}-\Delta t] \enspace .
\end{eqnarray*}
This allows to set the signal smoothly to zero and avoid creating discontinuities, as shown in \autoref{fig:TimeMask}.
The magnitude of this transformation is controlled by the length of masked signal $\Delta t$.}

\Cedric{The motivation of this augmentation is to teach the deep network to be less driven by isolated transient events.}
\Cedric{Indeed,}
as described in the \emph{AASM} scoring manual \cite{berry_aasm_2015}, sleep stages are most often characterized by the global information contained in a time window. 
For example, the sleep stage N1 is scored when more than 15 seconds ($\geq 50\%$) of a time window is dominated by theta activity ($4-7$\,Hz). 
The representations learned by the model should thus encapsulate the global information of the signal and avoid to rely on transient patterns. 
Consequently, one would expect two almost identical EEG windows differing only for a few seconds to be very close in the representation space \cite{cheng_subject-aware_2020}.
By masking part of the signal, we assign the same label (\eg sleep stage or action) to windows differing in the time domain only inside a short time span.

\begin{mdframed}[style=mystyle, frametitle=Time reverse]
    \Cedric{The \texttt{TimeReverse} augmentation (also noted \texttt{TRev}) was proposed in \cite{rommel2021cadda} and consists in randomly flipping the time axis in all channels.}
\end{mdframed}

\Cedric{More practically, we implement this augmentation by reversing the time indexing of the input signals $X$ with probability $\paug$:
\begin{equation*}
    \mathtt{TRev}[X](t) := \left\{
        \begin{array}{ll}
            X(t_{\max} - t) &\textrm{with } \paug, \\
            X(t) &\textrm{with } 1 - \paug,
        \end{array}
    \right.
\end{equation*}
where $t_{\max}$ is the length of a time window.}

\begin{figure*}[ht]
    \centering
    \includegraphics[width=0.9\textwidth]{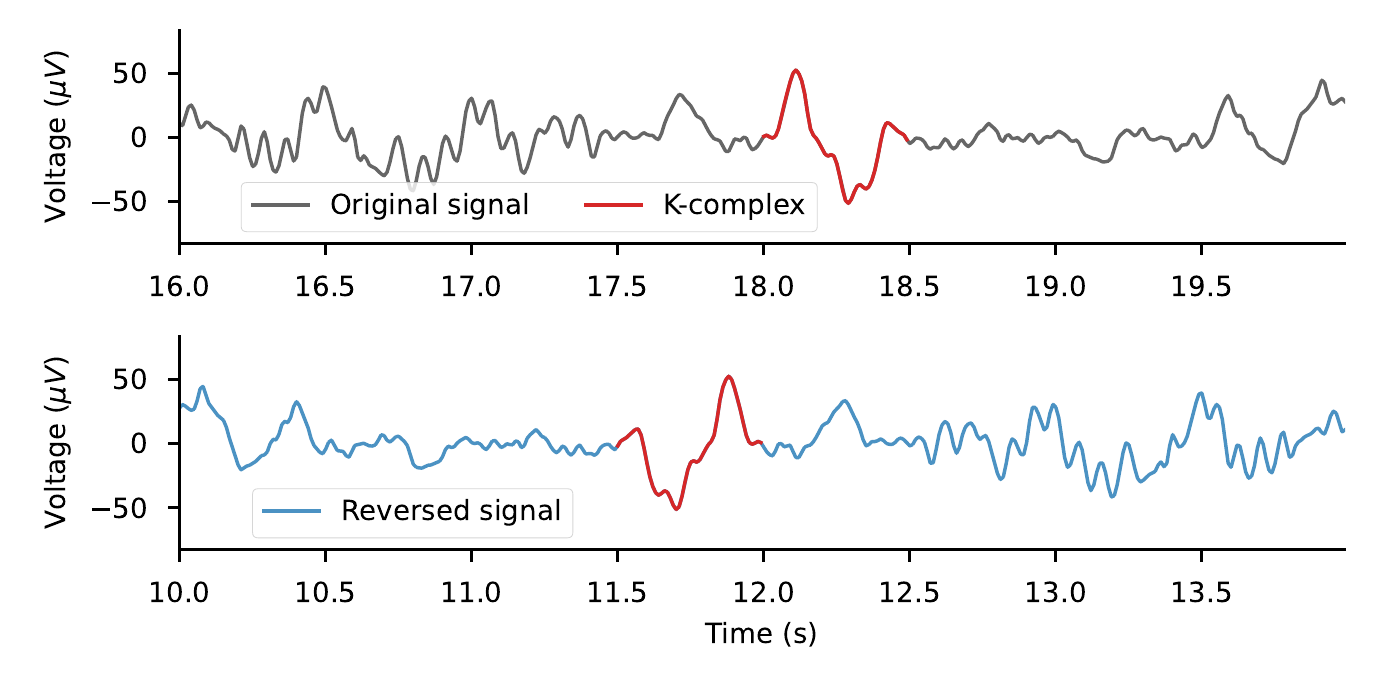}
    \caption{\texttt{TimeReverse} augmentation on an EEG signal from the \emph{SleepPhysionet} dataset. A large part of the signal is not deeply affected. Wave patterns are merely translated along the time axis.
    Some specific asymmetric EEG patterns such as K-complexes are reversed by this transformation.}
    \label{fig:TimeReverse}
\end{figure*}

\Cedric{The motivation for this data augmentation method is that the frequency power ratios contain a substantial part of the information useful for many EEG classification tasks.}
\Cedric{In sleep stage classification for example, some stages, such as N1, are scored based on the dominant rhythms observed (theta waves).}
Considering that the orientation of the time axis has no effect on the signal's power spectral densities (PSD),
\Cedric{it can be hypothesized that flipping the time axis generates a new input while preserving most of the information.}
%
%
Moreover, a large part of the time-domain information is also preserved by this transformation, since symmetric waveforms are merely shifted along the time axis and only asymmetric patterns are modified, as illustrated in \autoref{fig:TimeReverse}.
This
\Cedric{should}
be a useful property for instance to score the sleep stage N2,  which is characterized by more than two occurrences of K-complexes throughout the stage \cite{berry_aasm_2015}.

\begin{mdframed}[style=mystyle, frametitle=Sign flip]
    \Cedric{The \texttt{SignFlip} augmentation, introduced in \cite{rommel2021cadda}, consists in randomly inverting the sign of all EEG channels.}
\end{mdframed}

\Cedric{In practice, each new input signal $X$ is multiplied by $-1$ with probability $\paug$ (\lcf \autoref{subsec:EEG_tasks}):
\begin{equation*}
    \mathtt{SignFlip}[X](t) := \left\{
    \begin{array}{ll}
        -X(t) & \textrm{with } \paug,\\
        X(t) & \textrm{with } 1 - \paug.
    \end{array}
    \right.
\end{equation*}}

\Cedric{The motivation behind this augmentation is that it preserves the topographical properties of the electric field potential in terms of location and intensity, while changing its polarity.}
\Cedric{Indeed,}
the electric potentials measured with EEG are driven by post-synaptic potentials along dendrites of pyramidal neurons.
Depending on the geometric alignment of active neurons, the current produced can add up to be measured non-invasively with EEG. The group of neurons can be well modeled as electric current dipole, characterized by a moment vector at a given location in the brain.
For most analysis, the moment strength (norm) and location (origin) are sufficient.
While the moment direction is often unused, it
is responsible for the sign of the potential measured by the EEG device.
Our guess is that changing the sign of EEG channels will preserve the instrumental information contained in the strength and location of the dipole. In terms of physiology, it corresponds to current flowing from superficial cortical layers to deep layers or vice versa.


%

\subsection{\Cedric{Empirical comparison of time domain augmentations}}
\subsubsection{Parameters selection}

Here we investigate the effects of the transformations' magnitude on the classification performance.
As shown in \autoref{table:time_params}, the magnitude of \texttt{GaussianNoise} is controlled by its standard deviation $\sigma$, while \texttt{SmoothTimeMask} is governed by the length of the mask $\Delta t$.
Note that there is no notion of strength or magnitude for \texttt{TimeReverse} and \texttt{SignFlip}, which are hence not studied in this subsection.

\begin{figure}[ht]
     \centering
    \begin{subfigure}[b]{\columnwidth}
         \centering
         \includegraphics[width=\columnwidth]{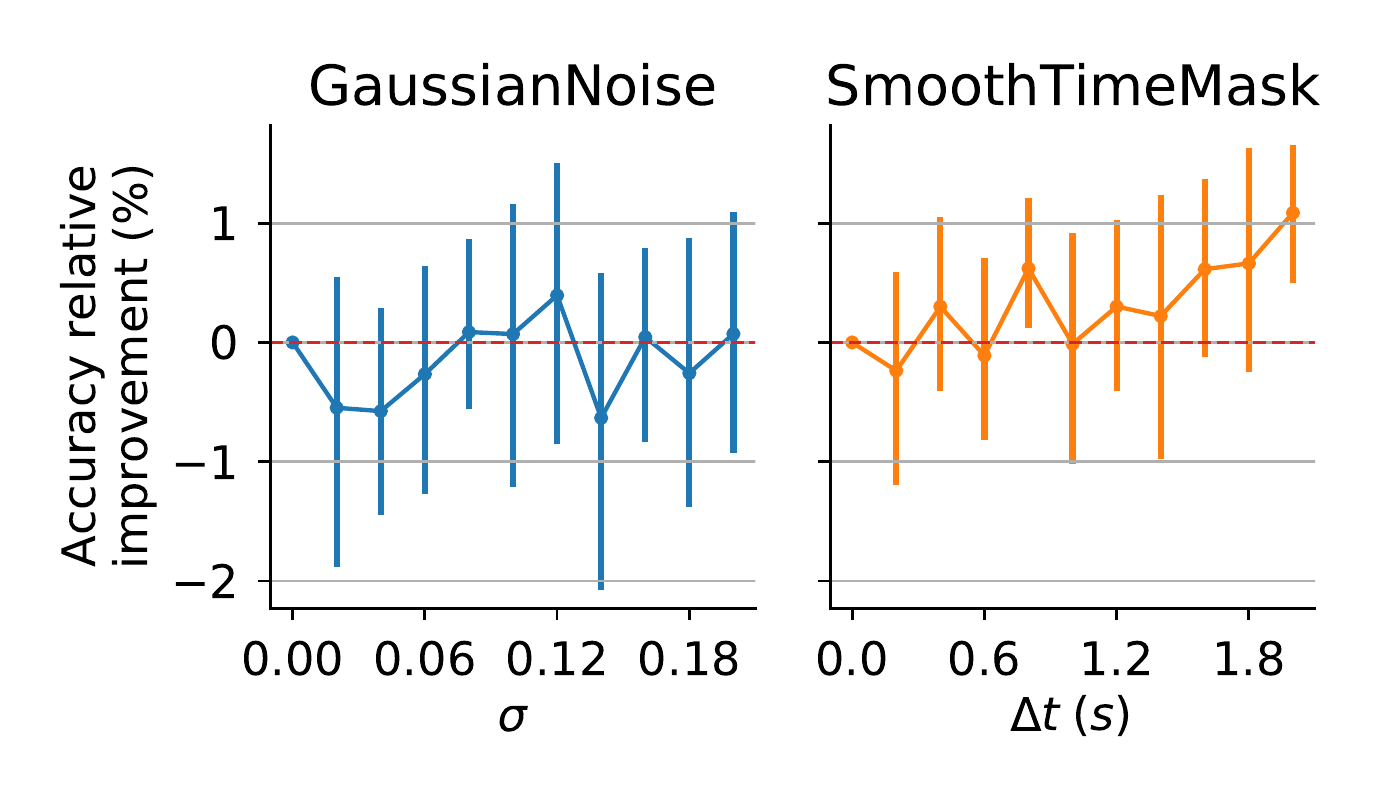}
         \caption{\emph{SleepPhysionet}}
         \label{fig:time_param_search_physionet}
     \end{subfigure}
    \begin{subfigure}[b]{\columnwidth}
         \centering
         \includegraphics[width=\columnwidth]{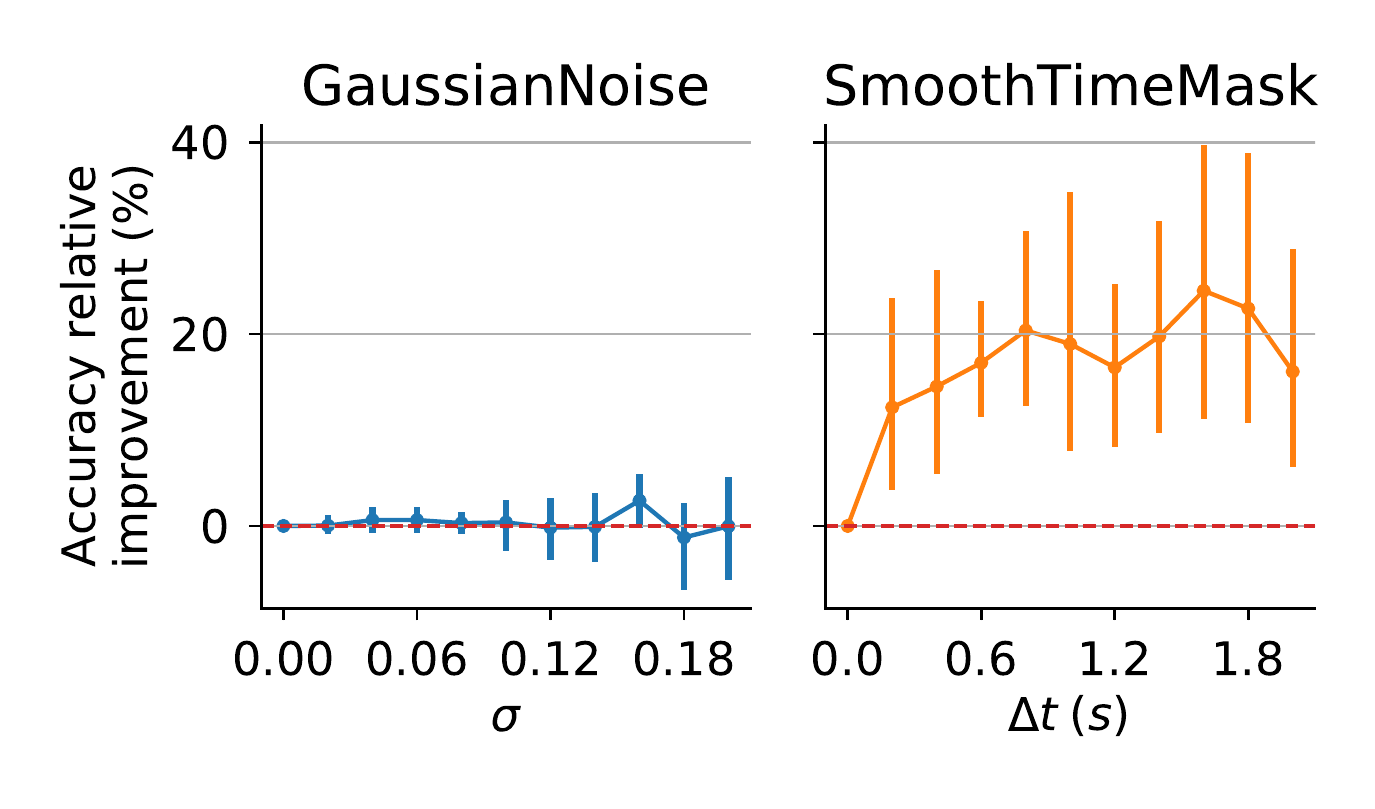}
         \caption{\emph{BCI IV 2a}}
         \label{fig:time_param_search_BCI}
     \end{subfigure}
     \caption{Time augmentations parameters selection on the \emph{SleepPhysionet} (a) and \emph{BCI IV 2a} (b) datasets. Models were trained on respectively 350 and 60 windows using augmentations parametrized with 10 different linearly spaced values.
     Validation accuracies are reported relatively to a model trained without data augmentation.
     The error bars correspond to the 95\% confidence intervals based on a 10-fold cross-validation.}
     \label{fig:time_param_search}
\end{figure}

\paragraph{SleepPhysionet}
%
As illustrated in \autoref{fig:time_param_search},
the impact of the magnitude is quite different between \texttt{SmoothTimeMask} and \texttt{GaussianNoise}.
While increasing the magnitude of \texttt{SmoothTimeMask} seems beneficial, no clear trend is observed for \texttt{GaussianNoise}.
For \texttt{SmoothTimeMask}, we restricted our experiment to masks of less than two seconds
to avoid removing too much crucial information from the signal.
It seems that the augmentation is more efficient with masks of maximum length.

\paragraph{BCI IV 2a}
The grid search on the \emph{BCI IV 2a} dataset presented in \autoref{fig:time_param_search_BCI} shares similarities with its counterpart on the \emph{SleepPhysionet} dataset (\autoref{fig:time_param_search_physionet}). 
In both cases, increasing the mask length enhances the performance of \texttt{SmoothTimeMask}, whereas \texttt{GaussianNoise} does not seem to lead to any robust improvement.
\texttt{SmoothTimeMask} appears to be much more useful on the BCI task though, since it produces relative improvements of up to $20\%$ when only $60$ windows per class are used for the training.
\begin{table*}[ht]
    \centering
    \small
    \begin{tabular}{c c c c c c} 
        \hline
        Augmentation & Parameter & Interval & Unit & Best value (sleep staging) & Best value (BCI)\\
        \hline
        \texttt{GaussianNoise} & $\sigma$ & $[0, 0.2]$ & - & 0.12 & 0.16\\
        \texttt{SmoothTimeMask} & $\Delta t$ & $[0, 2]$ & s & 2\,s & 1.6\,s\\
        \hline
    \end{tabular}
    \normalsize
    \caption{Adjustable parameter for each time domain augmentation.}
    \label{table:time_params}
\end{table*}

\subsubsection{Learning curves}
\label{sec:time-learning-curve}

\begin{figure*}[ht]
     \centering
     \begin{subfigure}[b]{\columnwidth}
         \centering
        \includegraphics[width=\columnwidth]{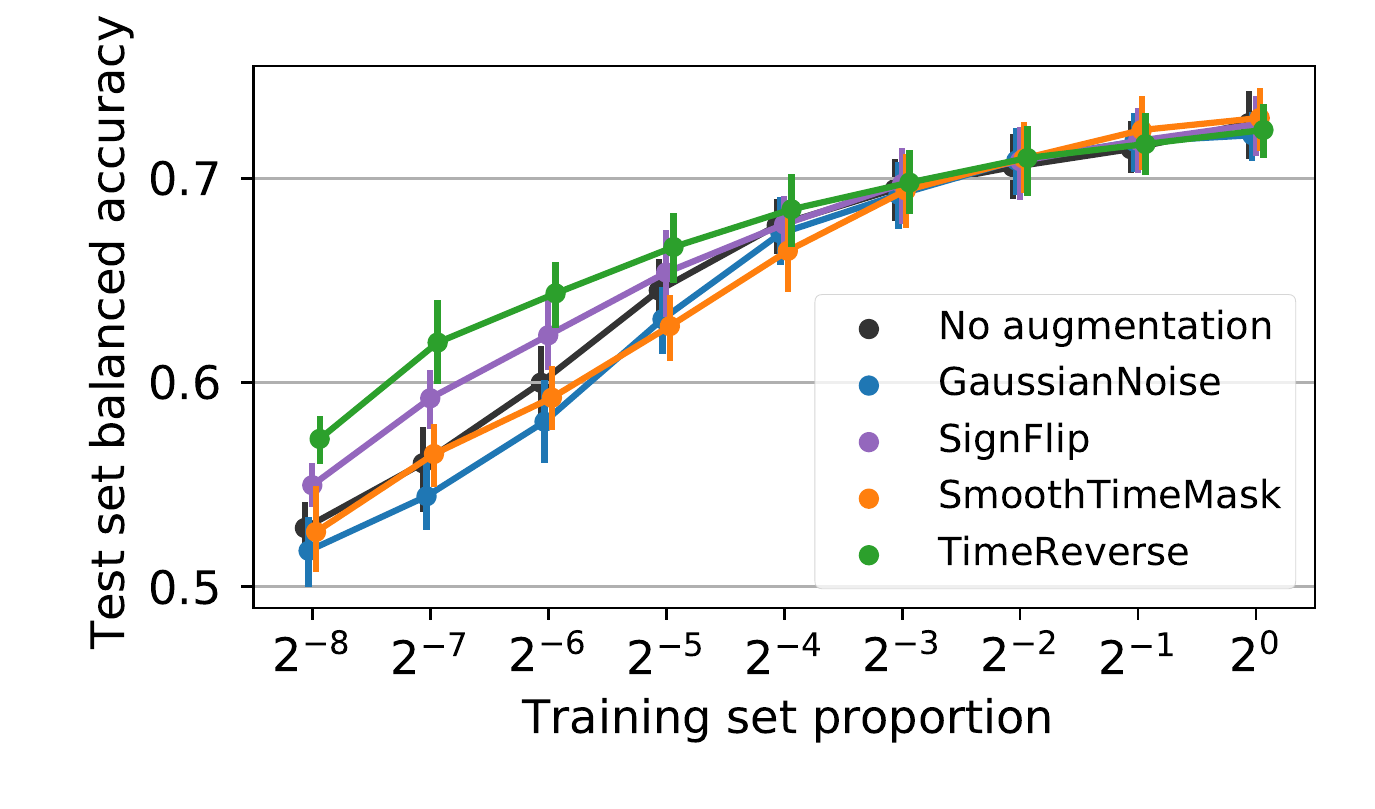}
         \caption{\emph{SleepPhysionet}}
         \label{fig:time_LRC_physionet}
     \end{subfigure}
     \begin{subfigure}[b]{\columnwidth}
         \centering
         \includegraphics[width=\columnwidth]{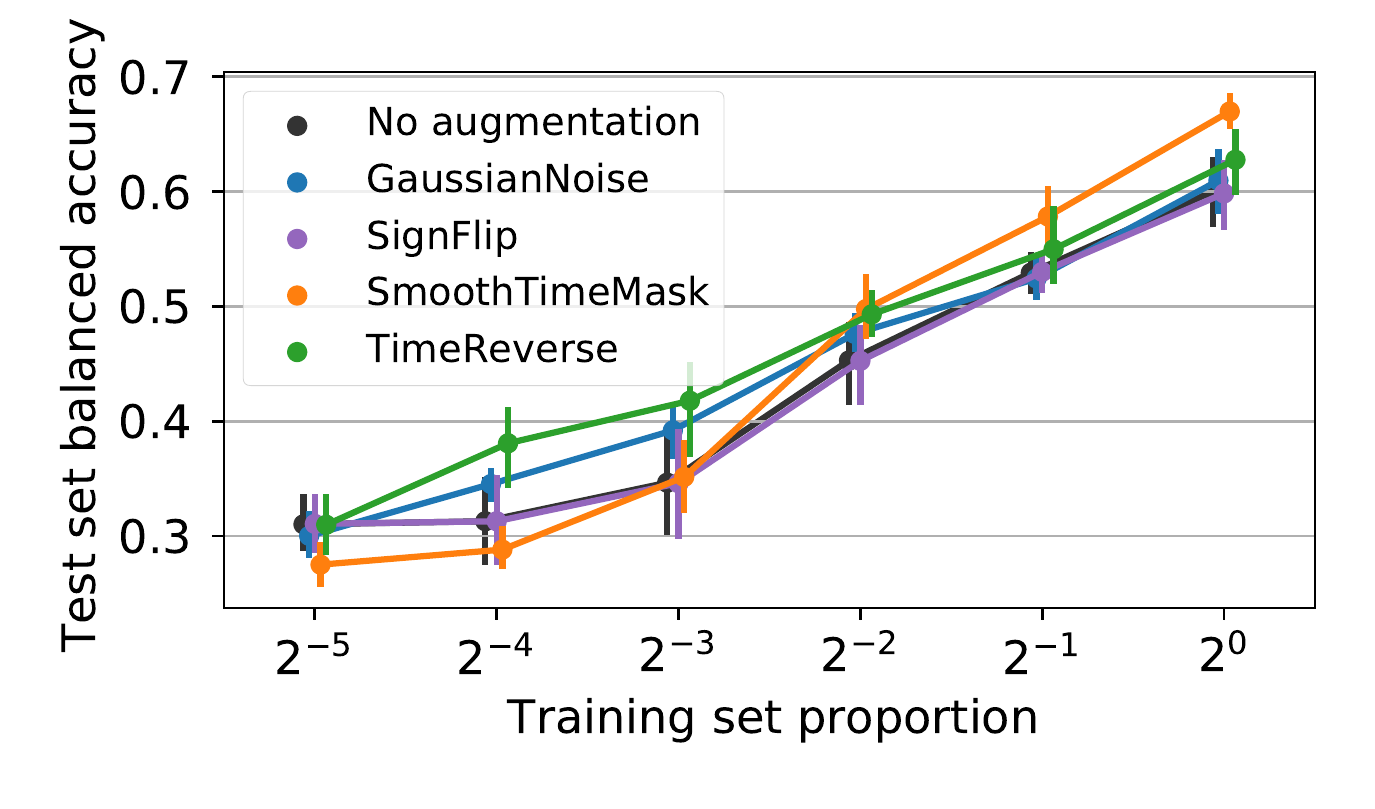}
         \caption{\emph{BCI IV 2a}}
         \label{fig:time_LRC_BCI}
     \end{subfigure}
     \caption{Learning curves for time domain augmentations along with the baseline trained with no augmentation. For each transformation, the same model is trained on fractions of the dataset of increasing size. After each training, the average balanced accuracy score on the test set is reported with error bars representing the 95\% confidence intervals estimated from 10-fold cross-validation.}
     \label{fig:time_LRC}
\end{figure*}

\paragraph{SleepPhysionet}
As expected from the previous experiment, \autoref{fig:time_LRC_physionet} confirms that \texttt{GaussianNoise} and \texttt{SmoothTimeMask} have a negligible effect on the sleep staging task and perform on par with the baseline model.
The most significant improvements are brought by \texttt{SignFlip} and \texttt{TimeReverse} with the latter outperforming the former.
This suggests that symmetries are notably relevant invariances for the sleep stage classification task. They preserve both the frequencies and some of the transient patterns occurring during polysomnographies, such as sleep spindles, which presents a symmetry both along the x and y axis.
This observation also supports the claim that sleep scoring heavily relies on frequency domain features since the two augmentations that leave the PSD of the signal unchanged outperform the others.

\paragraph{BCI IV 2a}
\autoref{fig:time_LRC_BCI} contains the learning curve plots for the \emph{BCI IV 2a} dataset.
It suggests that \texttt{TimeReverse} and \texttt{SmoothTimeMask} are the most suited time domain augmentations for the BCI task in low and high data regimes respectively.
An intuitive interpretation of this result is that, in motor imagery, subjects are asked to mentally simulate the same physical action during the whole trial and the information encoded in brain signals is hence invariant to local time distortions.
Another striking observation has to do with the learning curve of \texttt{SignFlip}, which is consistently equivalent to the baseline's
while it is the second most efficient augmentation in \emph{SleepPhysionet}.
This happens because, unlike in the sleep staging experiments, the network architecture used here includes a layer which squares the activations, thus already encoding the sign invariance.

\subsubsection{Per class analysis}
\label{subsubsec:time_per_class}

\paragraph{SleepPhysionet}

\begin{figure*}[ht]
     \centering
     \begin{subfigure}[b]{\columnwidth}
         \centering
         \includegraphics[width=\columnwidth]{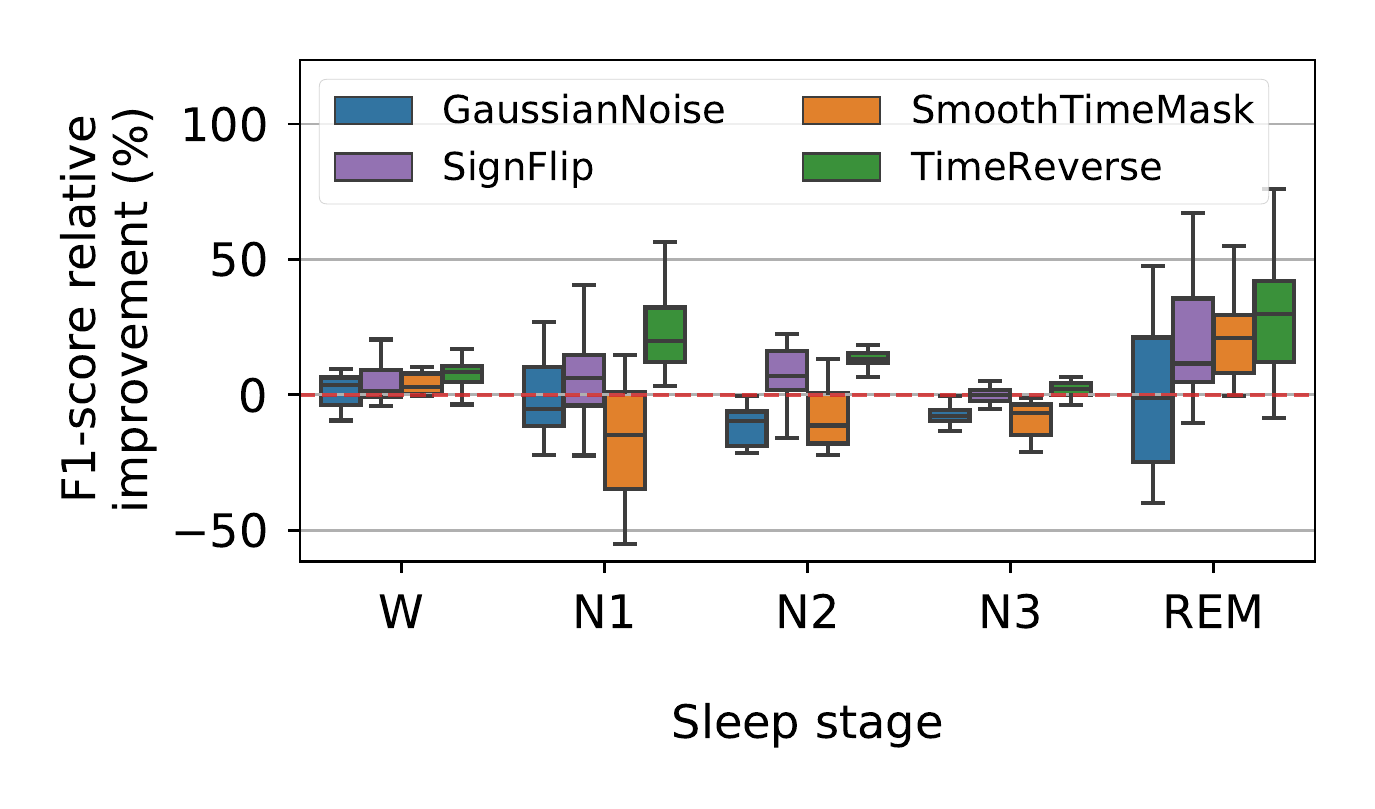}
         \caption{\emph{SleepPhysionet}}
         \label{fig:time_boxplot_physionet}
     \end{subfigure}
     \begin{subfigure}[b]{\columnwidth}
         \centering
         \includegraphics[width=\columnwidth]{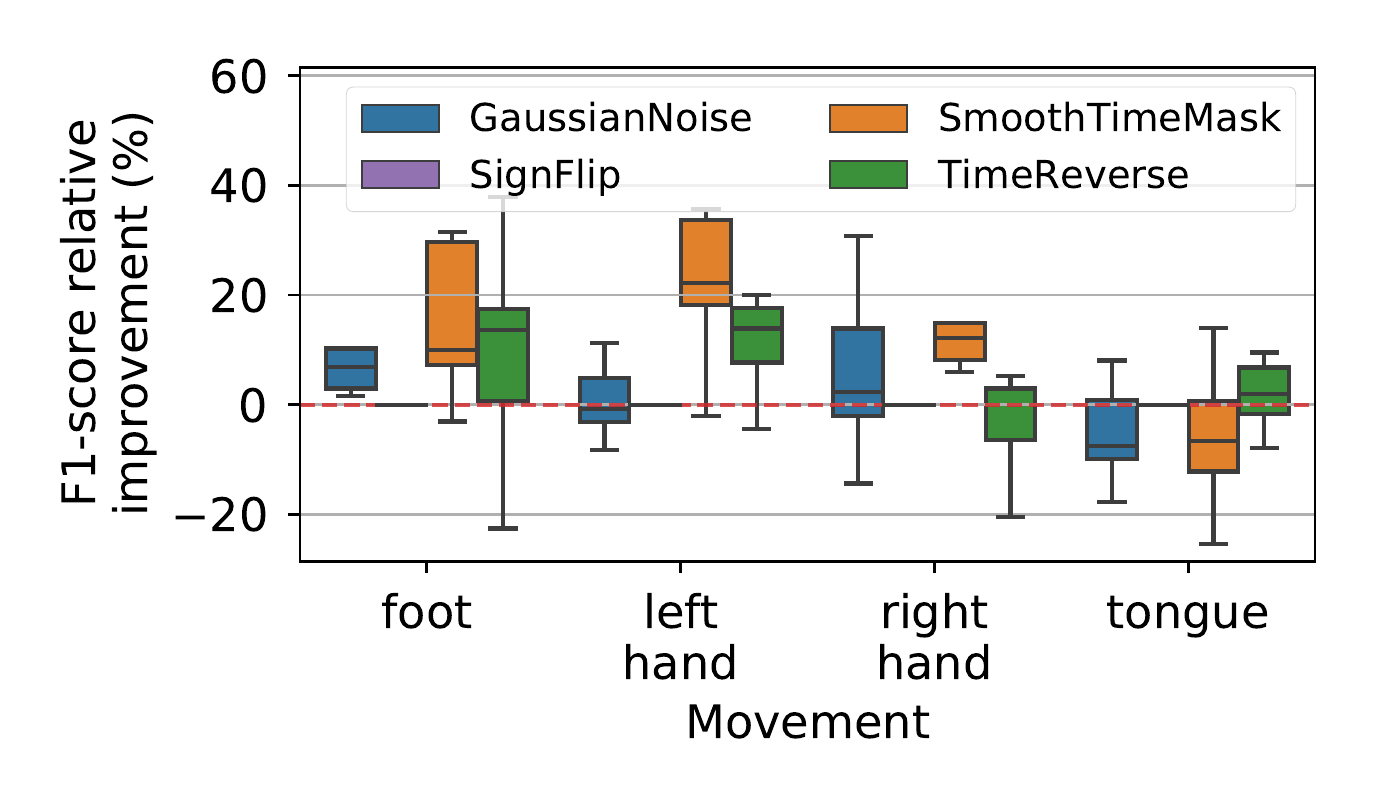}
         \caption{\emph{BCI IV 2a}}
         \label{fig:time_boxplot_BCI}
     \end{subfigure}
     \caption{Per-class F1-score for time domain transformations. Scores are reported as relative improvement over a baseline trained without data augmentation. Models were trained on $180$ and $230$ time windows for \emph{SleepPhysionet} and \emph{BCI IV 2a} datasets respectively. Boxplots were estimated using $10$-fold cross-validation.}
\end{figure*}

\autoref{fig:time_boxplot_physionet} introduces the results per class of the time domain augmentations for the sleep staging task.
The REM stage benefits more than the others from this class of transformations.
This stage is characterized by low amplitude mixed frequency and bursts of eye activity.
The scoring thus relies heavily on global signal amplitudes which are mostly preserved with time domain augmentations (except for \texttt{GaussianNoise} which performs the worst among them).
\texttt{TimeReverse} especially yields the best results for this class as well as for all others.
Note also that while \texttt{SmoothTimeMask} leads to improvements for REM and W stages, it appears to be detrimental for learning to recognize non-REM sleep.
A possible interpretation of this observation is that this transformation may erase important waves, such as K-complexes and spindles, which strongly characterize stages N2 and N3.

\paragraph{BCI IV 2a}

\autoref{fig:time_boxplot_BCI} introduces the results per class of the time domain augmentations for the BCI motor imagery task.
Here again, we observe that \texttt{SignFlip} is equivalent to the baseline with no augmentation, as explained in \autoref{sec:time-learning-curve}.
While \texttt{GaussianNoise} also does not help for any imagined movement, \texttt{TimeReverse} and \texttt{SmoothTimeMask} greatly improve the predictive accuracy for most classes, specially left-hand, right-hand and foot.
%

\subsection{\Cedric{Conclusion of time augmentations experiments}}

\Cedric{The \texttt{TimeReverse} augmentation consistently yielded the largest performance improvements with small training sets relative to the baseline trained without data augmentation: up to $+13\%$ for sleep stage classification and $+25\%$ for motor imagery experiments\footnote{At training set fractions of $2^{-7}$ and $2^{-4}$ respectively.}.}
\Cedric{With larger training sets, \texttt{SmoothTimeMask} seems like the best data augmentation for motor imagery, while it appears to be detrimental in sleep staging.}
\Cedric{The same can be observed for \texttt{GaussianNoise}, although it does not bring the same level of improvements as \texttt{SmoothTimeMask} in our BCI experiments.}

\section{Frequency domain augmentations}
\label{sec:frequency-tfs}

\subsection{\Cedric{Rationale of the frequency augmentations}}

\begin{mdframed}[style=mystyle,frametitle=Frequency shift]
    \Cedric{The \texttt{FrequencyShift} augmentation, proposed in \cite{rommel2021cadda}, consists in shifting the spectrum of the signals of all EEG channels by a random frequency $\Delta f$.}
\end{mdframed}

\Cedric{In practice the shift is performed on the complex analytic signal associated to the EEG signal $X$, ${X_a = X + j\mathcal{H}(X)}$, where $\mathcal{H}$ denotes the Hilbert transform:}
\Cedric{
\begin{equation*}
    \mathtt{FrequencyShift}[X](t) := \textrm{Re}(X_a(t) \cdot e^{2i\pi \Delta f\cdot t})
\end{equation*}
Here one takes the real part to recover the shifted signal.
The shift value $\Delta f$ is randomly sampled with uniform probability in an interval ${[-\Delta f_{\max}, +\Delta f_{\max}]}$ each time an EEG window is augmented. The parameter $\Delta f_{\max}$ is used to set the magnitude of this transformation.}

\begin{figure*}[ht]
    \centering
    \includegraphics[width=0.9\textwidth]{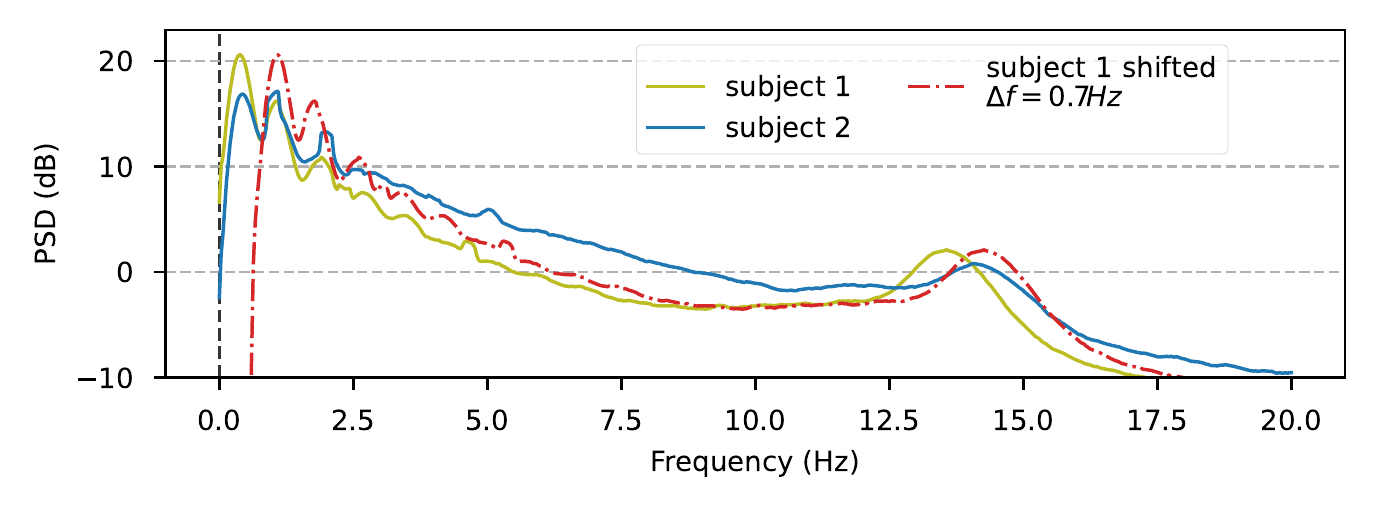}
    \caption{Averaged power spectral density of windows corresponding to the sleep stage N2, for two different subjects from the \emph{SleepPhysionet} dataset. The red dash-dot curve corresponds to the recording of subject 1 transformed using the \texttt{FrequencyShift} augmentation.
    It allows to translate the PSD peak close to subject 2.}
    \label{fig:PSD-shift}
\end{figure*}

\Cedric{The motivation for this data augmentation method is that a substantial part of the information useful for many EEG classification tasks lies in the frequency domain of the signal.}
In sleep scoring for example, most sleep stages are characterized by the occurrence of specific brain rhythms in a given frequency range.
The stage N2 can be characterized by so-called sleep spindles with frequencies located between $12$ and $15$\,Hz, while stage N3 is characterized by slow waves in the delta band \cite{berry_aasm_2015}.
The predominance of certain brain rhythms is well captured by the power spectral density (PSD) of the EEG signals, which 
\Cedric{have peaks at specific frequency ranges,}
as depicted in \autoref{fig:PSD-shift}.
Due to strong inter-subject variability, the location of these peaks
\Cedric{for a given sleep stage}
can be slightly different from one subject to another,
\Cedric{and the \texttt{FrequencyShift} augmentation aims to mimic this variability.}
%

\begin{mdframed}[style=mystyle, frametitle=Fourier transform surrogate]
    \Cedric{The Fourier Transform surrogate augmentation, proposed in \cite{schwabedal_addressing_2019} and denoted \texttt{FTSurrogate} hereafter, consists in randomizing the phase of the Fourier coefficients of EEG signals.}
\end{mdframed}

\Cedric{In practice, this transformation is performed by
\Cedric{computing the Fourier coefficients of all EEG channels and}
adding random noise to their phase:
\[ 
\mathcal{F}[\mathtt{FTSurrogate}(X)](f)= \mathcal{F}[X](f)e^{i\Delta\varphi} \enspace,
\] 
where $\mathcal{F}$ is the Fourier transform operator,
$f$ is a frequency and $\Delta\varphi$ is a frequency-specific random phase perturbation. The augmented signals are obtained with the inverse Fourier transform.}
In our implementation, the values of $\Delta\varphi$ \Cedric{for each frequency $f$} are uniformly sampled in an interval $[0, \Delta\phimax]$,
where $\Delta\phimax\in[0, 2\pi)$ controls the magnitude of the transformation.
\Cedric{This transformation can be independently applied to each channel, or by enforcing that the phase shift is common for all channels. While EEG channels are perturbed independently in our sleep classification experiments, we found that perturbing all channels equally is crucial in BCI experiments to preserve cross-channel correlations and obtain good results.}

\begin{figure*}[ht]
    \centering
    \includegraphics[width=0.9\textwidth]{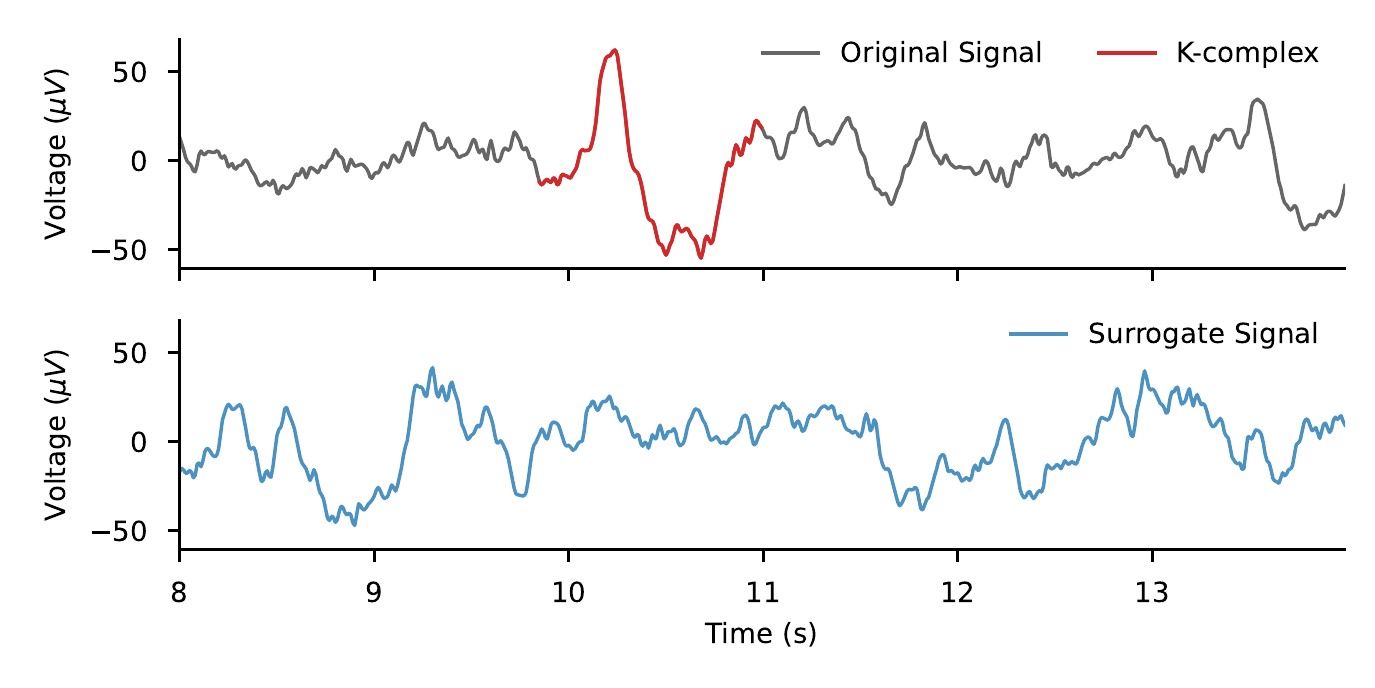}
    \caption{Effect of \texttt{FTSurrogate} on transient patterns.
    The original extract from a window scored as N2 presents a highly localized K-complex, whereas the surrogate does not.}
    \label{fig:FTSurrogate_kcomplex}
\end{figure*}

\Cedric{The motivation behind this augmentation is that it preserves the frequency-bands power ratios, yet it leads to a global change of the signal’s representation in the time-domain.}
\Cedric{This is illustrated on \autoref{fig:FTSurrogate_kcomplex}, which shows that characteristic patterns \Cedric{in the time-domain} such as K-complexes are erased by this operation.}
As a result, it decreases the model's reliance on the signal's \Cedric{waveform,}
encouraging the model to
\Cedric{leverage} the PSD \Cedric{information}.
%

The Fourier transform surrogate method is based
on the assumption
that EEG signals are well described by stationary linear stochastic processes \cite{schwabedal_addressing_2019}.
As such, they must be uniquely characterized by the amplitudes of their Fourier coefficients and must have random phases in $[0, 2\pi)$.
\Cedric{This implies that each frequency in the signal is considered independent,}
which might \Cedric{appear as a}
strong assumption. Indeed, neural signals contain transient events, such as K-complexes,
which spread across multiple frequencies.
\Cedric{This creates}
some dependence among the phases of multiple Fourier coefficients (cf. \autoref{fig:FTSurrogate_kcomplex}). Besides transient events, phenomena known as cross-frequency coupling (CFC) have been reported in human electrophysiology signals~\cite{canolty2006,duprelatour2017nonlinear}.
Nevertheless, the hypothesis of stationary linear stochastic processes have been shown to be compatible with EEG signals in \cite{andrzejak_indications_2001}.
Plausible arguments in favor of this hypothesis are the huge number of neurons included in an EEG recording, the complicated structure of the brain and the possible blur of dynamical structures due to the different conductivities of the skull and other intermediate tissues.
Thus, while the Fourier transform surrogate method might be useful on the context of EEG, it is likely to be less relevant for intracranial recordings,
\Cedric{which have a better signal-to-noise ratio.}

\begin{mdframed}[style=mystyle, frametitle=Band-stop filter]
    \Cedric{The \texttt{BandstopFilter} augmentation, proposed for example in \cite{cheng_subject-aware_2020, mohsenvand_contrastive_2020}, consists in filtering out from all EEG channels a given frequency band selected at random.}
\end{mdframed}

\Cedric{Our implementation of this augmentation uses the finite impulse response notch filter from \textit{MNE-Python} \cite{gramfort_meg_2013}.
For each augmented EEG signal, the center of the frequency band is randomly picked from a uniform distribution between $0$ and $38$\,Hz, which is the approximate low-pass filtering frequency used to preprocess the datasets.
The width of the filter allows to control the magnitude of this transform.}

\Cedric{The motivation of this augmentation is to prevent machine learning models from overfitting on subject specific features and from relying too much on a few narrow frequency regions.}
\Cedric{Indeed,}
some patterns in EEG signals, such as the K-complex, have frequency signatures that spread over multiple frequency bands.
\Cedric{By filtering randomly selected regions of the PSD,}
one can promote models that use the full spectral information of these patterns.
%
%
\Cedric{In a sense,}
this transformation can be compared to \Cedric{the commonly used} dropout layer \cite{srivastava_dropout_2014},
\Cedric{which prevents artificial neural networks from relying too much on certain neurons.}

\subsection{\Cedric{Empirical comparison of frequency augmentations}}

\subsubsection{Parameters selection}
\label{subsubsec:freq_param_selection}

The parameters to be set for frequency domain augmentations are listed in \autoref{table:freq_params}.

\paragraph{\textit{SleepPhysionet}}

\begin{figure*}[ht]
     \centering
     \begin{subfigure}[b]{1.\textwidth}
         \centering
         \includegraphics[width=\textwidth]{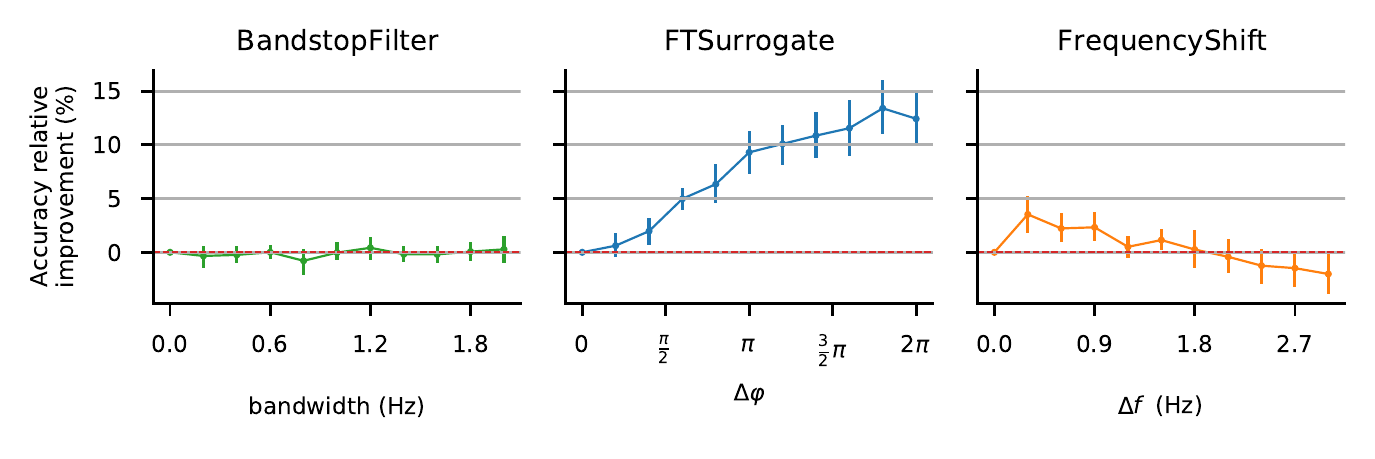}
         \caption{\emph{SleepPhysionet}}
         \label{fig:freq_param_search_physionet}
     \end{subfigure}
     \begin{subfigure}[b]{1.\textwidth}
         \centering
         \includegraphics[width=\textwidth]{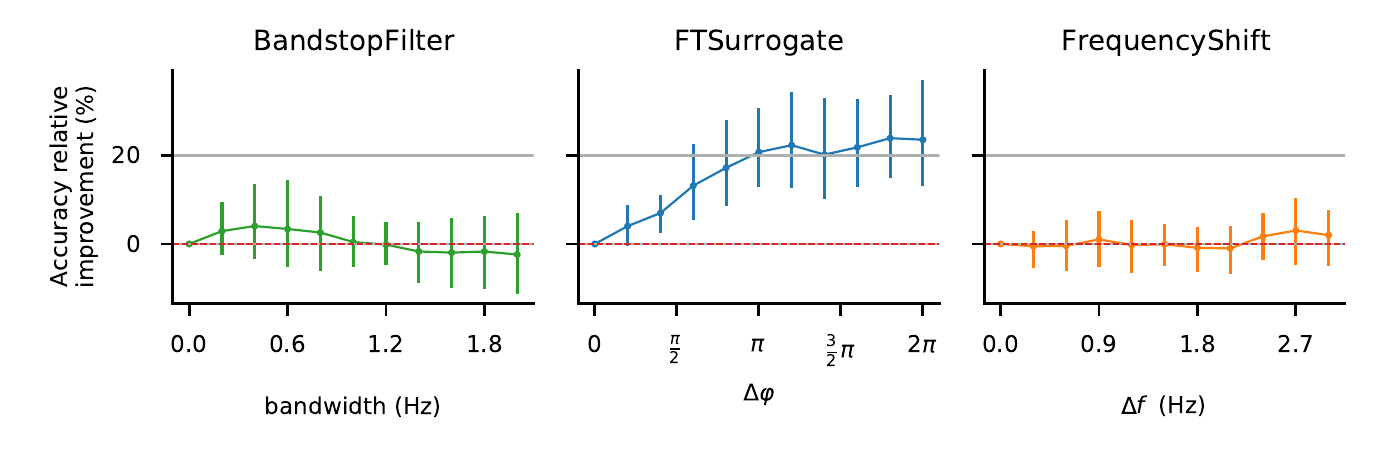}
         \caption{\emph{BCI IV 2a}}
         \label{fig:freq_param_search_BCI}
     \end{subfigure}
     \caption{Frequency augmentations parameters selection on the \emph{SleepPhysionet} (a) and \emph{BCI IV 2a} (b) datasets. Models were trained on respectively 350 and 60 windows using augmentations parametrized with 10 different linearly spaced values.
     Validation accuracies are reported relatively to a model trained without data augmentation.
     The error bars correspond to the 95\% confidence intervals based on a 10-fold cross-validation.}
     \label{fig:freq_param_search}
\end{figure*}

The results on the sleep staging task presented in \autoref{fig:freq_param_search} and \autoref{table:freq_params} reveal that the optimal magnitude varies significantly from one transformation to the other. 
The \texttt{FTSurrogate} augmentation benefits from the larger possible range of $\Delta \varphi$ values, corresponding to a nearly completely random phase selection within the interval $[0, 2\pi)$. 
On the contrary, \texttt{FrequencyShift} works better for small $\Delta f_{\max}$ values. 
An interpretation for this result could be that small frequency shifts allow to capture the inter-subject variability whereas larger values mix-up the frequency bands characterizing different classes.
Finally, \texttt{BandstopFilter} shows marginally better results for a bandwidth of $1.2$\,Hz, although performance gains compared to the baseline are not significant.

\paragraph{BCI IV 2a}

The results for \emph{BCI IV 2a} presented in \autoref{fig:freq_param_search_BCI} \Cedric{reveal} several differences compared to \emph{SleepPhysionet}.
Unlike for the sleep stage classification task, smaller bandwidths seem to work better for \texttt{BandstopFilter}.
However, 
this transformation does not lead to statistically significant improvements here again.
Maximum frequency shifts $\Delta f_{\max}$ privileged for 
\Cedric{the BCI}
task are
\Cedric{also}
higher than for sleep staging.
Finally, the \texttt{FTSurrogate} augmentation shows a pattern similar to the sleep staging task, as it benefits from higher $\Delta\varphi_{\max}$ \Cedric{values}.
\begin{table*}[ht]
    \centering
    \small
    \begin{tabular}{c c c c c c} 
        \hline
        Augmentation & Parameter & Interval & Unit & Best value (sleep staging) & Best value (BCI)\\ 
        \hline
        \texttt{BandstopFilter} & bandwidth & $[0, 2]$ & Hz & $1.2$\,Hz & $0.4$\,Hz\\
        \texttt{FTSurrogate} & $\Delta \phimax$ & $[0, 2\pi)$ & rad & $\frac{9}{10}\pi$ & $\frac{9}{10}\pi$\\
        \texttt{FrequencyShift} & $\Delta f_{\max}$ & $[0, 3]$ & Hz & $0.3$\,Hz & $2.7$\,Hz\\
        \hline
    \end{tabular}
    \normalsize
    \caption{Adjustable parameter of each frequency domain augmentation.}
    \label{table:freq_params}
\end{table*}

\subsubsection{Learning curves}
\label{sec:freq-learning-curves}
\begin{figure*}[ht]
    \centering
    \begin{subfigure}[b]{\columnwidth}
    \centering
        \includegraphics[width=\columnwidth]{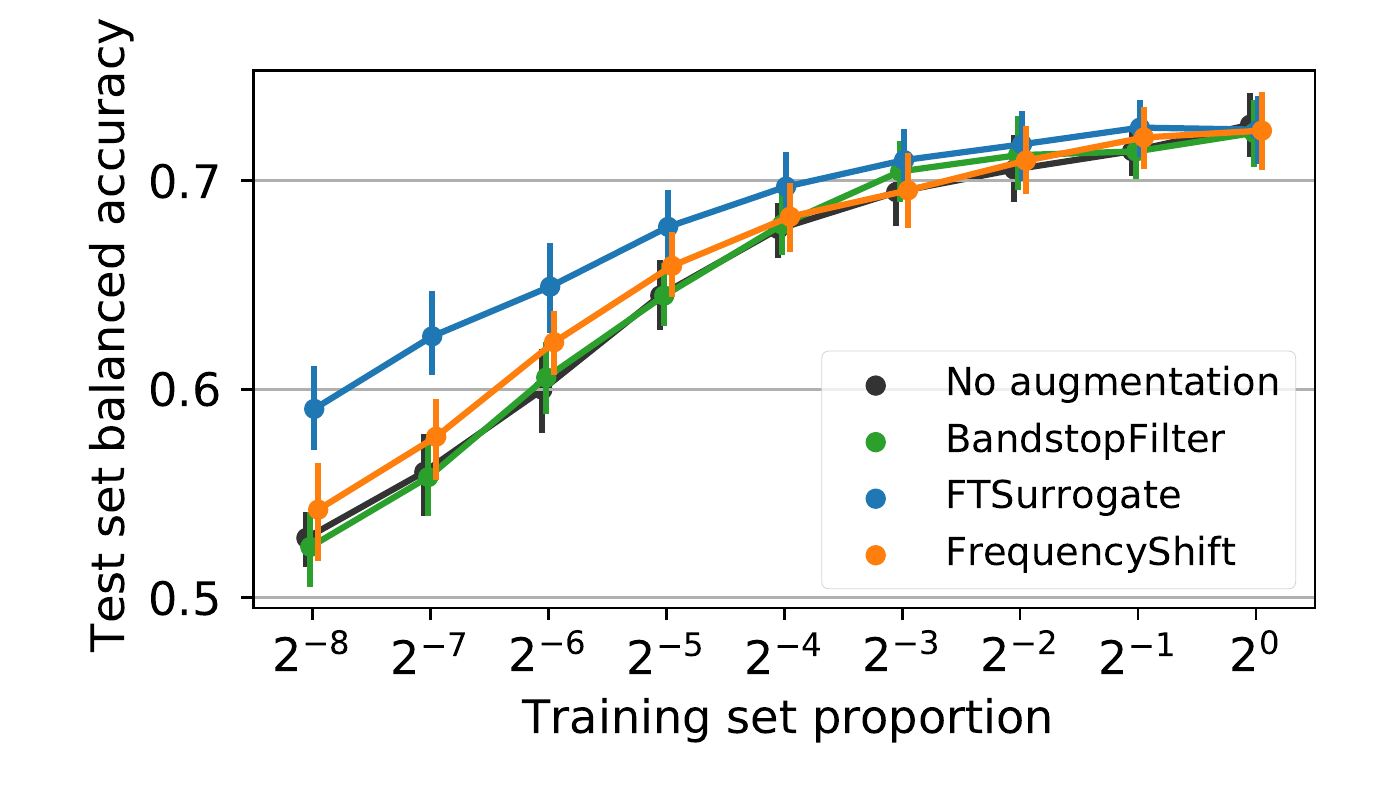}
        \caption{\emph{SleepPhysionet}}
        \label{fig:freq_LRC_physionet}
    \end{subfigure}
    \hfill
    \begin{subfigure}[b]{\columnwidth}
        \centering
        \includegraphics[width=\columnwidth]{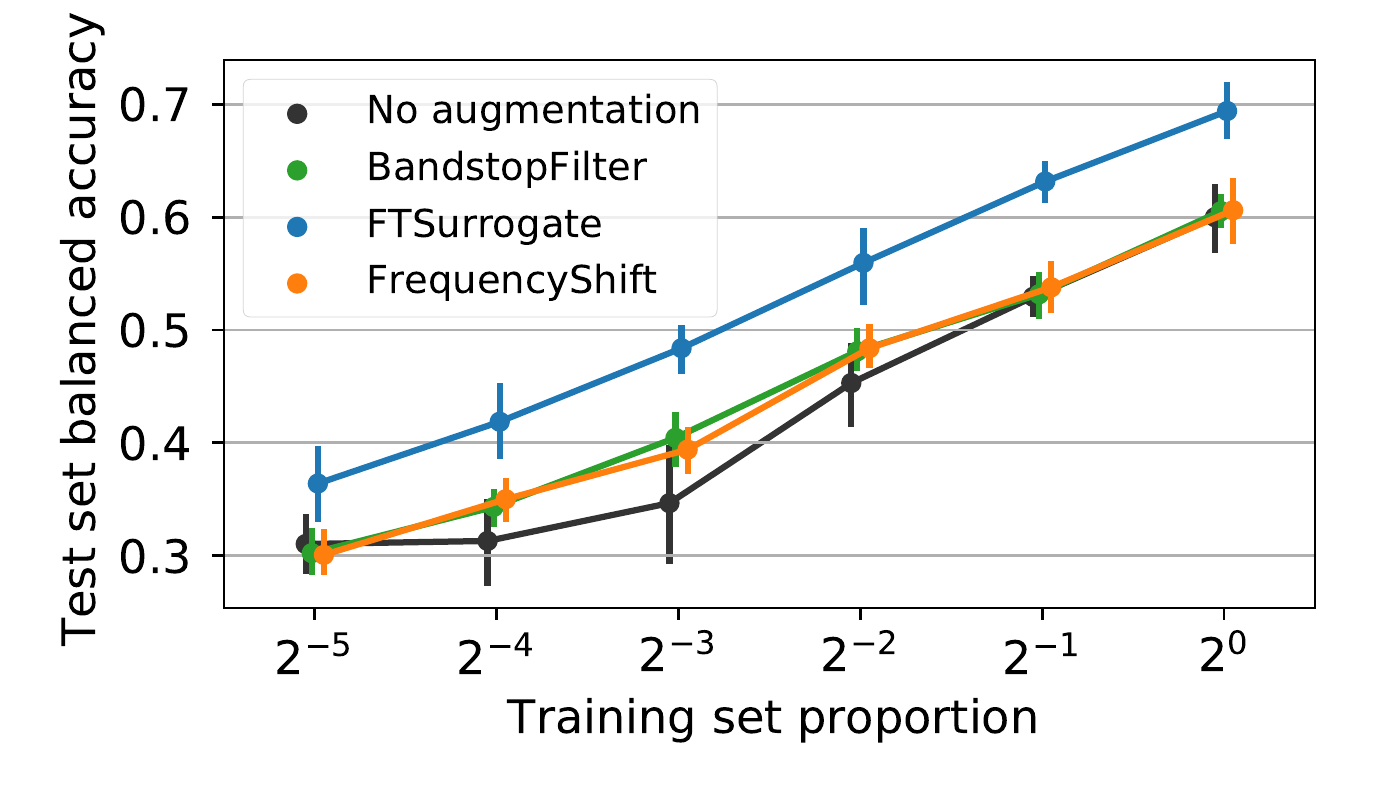}
        \caption{\emph{BCI IV 2a}}
        \label{fig:freq_LRC_BCI}
    \end{subfigure}
    \caption{Learning curves for frequency domain augmentations along with the baseline trained with no augmentation. For each transformation, the same model is trained on 8 fractions of the dataset of increasing size. After each training, the average balanced accuracy score on the test set is reported with error bars representing the 95\% confidence intervals estimated from 10-fold cross-validation.}
    \label{fig:freq_LRC}
\end{figure*}

\paragraph{SleepPhysionet}
As expected, the first learning curve experiment depicted in \autoref{fig:freq_LRC_physionet} reveals that data augmentation methods are more helpful in low data regimes. 
It helps to mitigate the lack of data by artificially increasing the training set size.
For example, a model trained with \texttt{FTSurrogate} on a
\Cedric{small fraction of training data ($2^{-4}$)}
achieves performances comparable to a model trained without augmentation on \Cedric{four times} as many data points \Cedric{($2^{-2}$)}.
\Cedric{Moreover, this augmentation appears as the top performer in this task, yielding balanced accuracy relative gains of up to $12\%$ in low data regimes compared to the baseline.}
\Cedric{While \texttt{FrequencyShift} also brings some smaller performance improvements, \texttt{BandstopFilter} is ineffective on this dataset.}
\Cedric{This is evidence that preserving the frequencies' power ratios is important in sleep stage classification.}

\paragraph{BCI IV 2a}
As in the sleep stage classification task, \texttt{FTSurrogate} outperforms other frequency domain augmentations on the \emph{BCI IV 2a} dataset, as shown in \autoref{fig:freq_LRC_BCI}.
\Cedric{Namely, a model trained with \texttt{FTSurrogate} on half of the training set reaches a test accuracy higher than the same model trained without data augmentation on the whole training set.}
\Cedric{It also leads to a 45\% relative improvement in accuracy compared to the model trained without augmentation on a training subset $2^{-3}$ times smaller than the original one.}
\Cedric{Moreover, this augmentation seems to yield significant improvements over all training set sizes considered in this experiments, unlike in sleep stage classification where it mostly helped with smaller training sets.}

\Cedric{Another similarity with the sleep stage classification task are the good results of \texttt{FrequencyShift},}
\Cedric{whose performance improvements even exceed those observed on the \emph{SleepPhysionet} dataset.}
This is surprising given that this transformation was originally designed to simulate the inter-subject variability observed in sleep stages.
Despite these similarities, a major difference between sleep stage classification and BCI is that \texttt{BandstopFilter} appears as a relevant data augmentation technique for the latter, since it is shown to improve accuracy by up to 17\% in low data regimes.
Overall, these results suggest that a critical part of the information lie in the frequency domain for both classification tasks.

\subsubsection{Per class analysis}
\label{sec:freq-class-wise}
\begin{figure*}[ht]
    \centering
    \begin{subfigure}[b]{\columnwidth}
    \centering
        \includegraphics[width=\columnwidth]{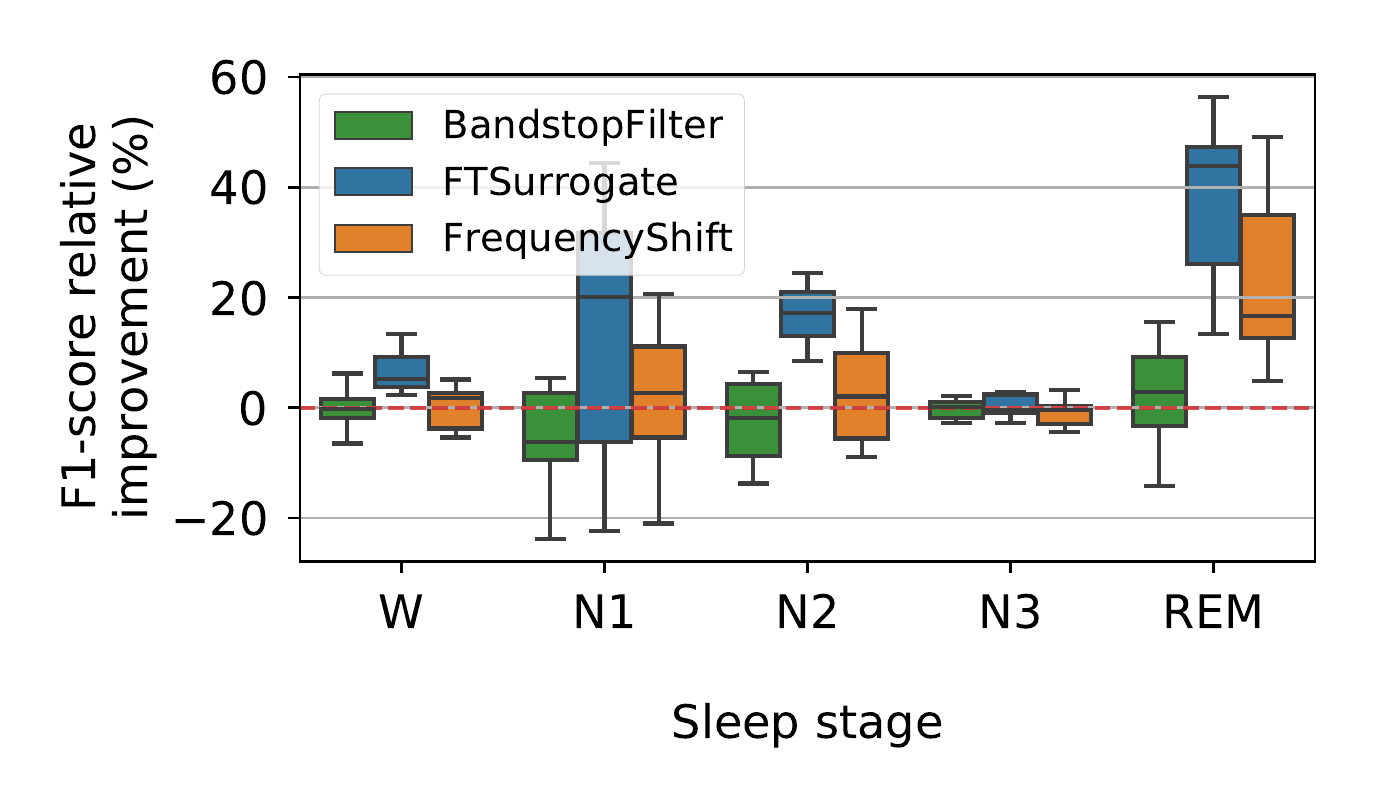}
        \caption{\emph{SleepPhysionet}}
        \label{fig:freq_boxplot_physionet}
    \end{subfigure}
    \hfill
    \begin{subfigure}[b]{\columnwidth}
        \centering
            \includegraphics[width=\columnwidth]{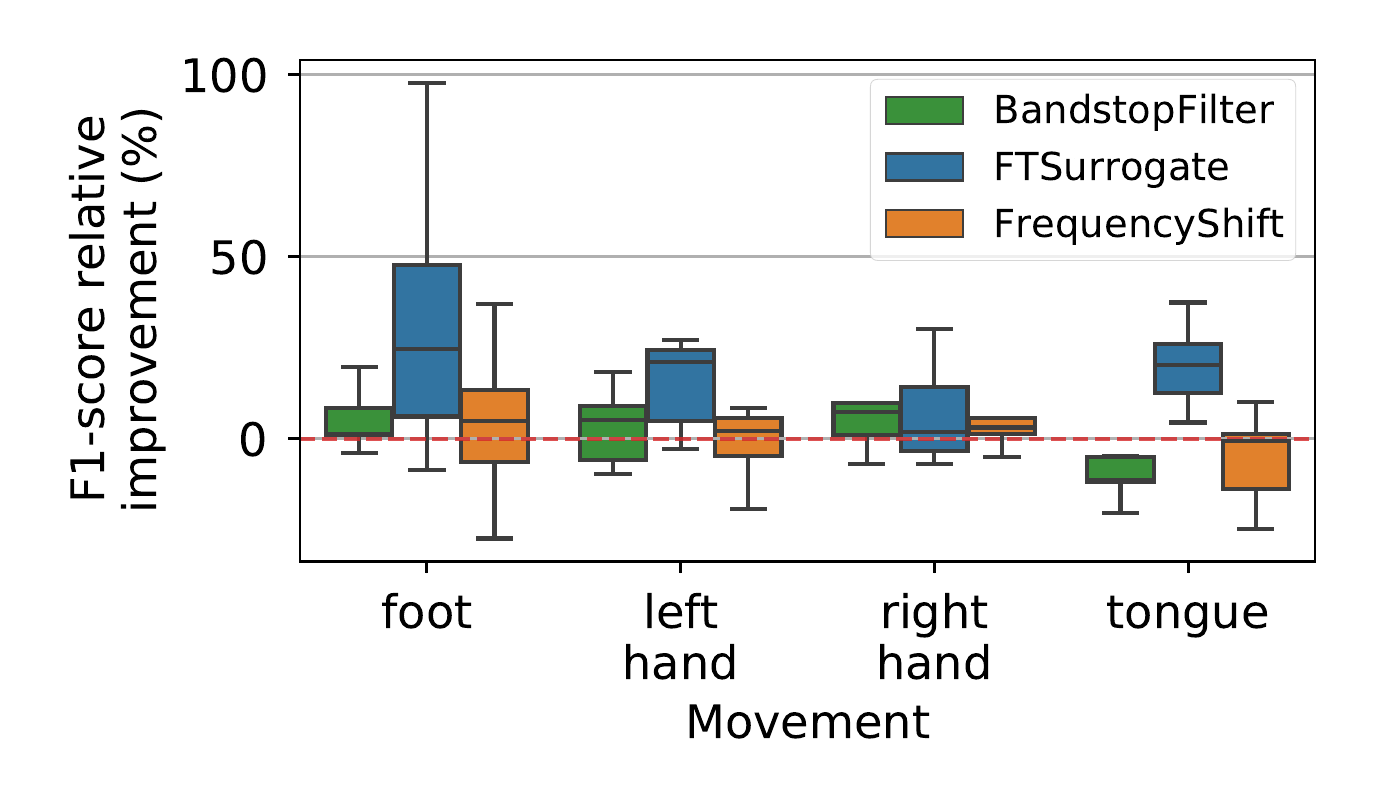}
        \caption{\emph{BCI IV 2a}}
        \label{fig:freq_boxplot_BCI}
    \end{subfigure}
    \caption{Per-class F1-score for frequency domain transformations. Scores are reported as relative improvement over a baseline trained without data augmentation. Models were trained on $180$ and $230$ time windows for \emph{SleepPhysionet} and \emph{BCI IV 2a} datasets respectively. Boxplots were estimated using $10$-fold cross-validation.}
\end{figure*}

\begin{figure*}[ht]
    \centering
    \begin{subfigure}[b]{\columnwidth}
        \centering
        \includegraphics[width=\columnwidth]{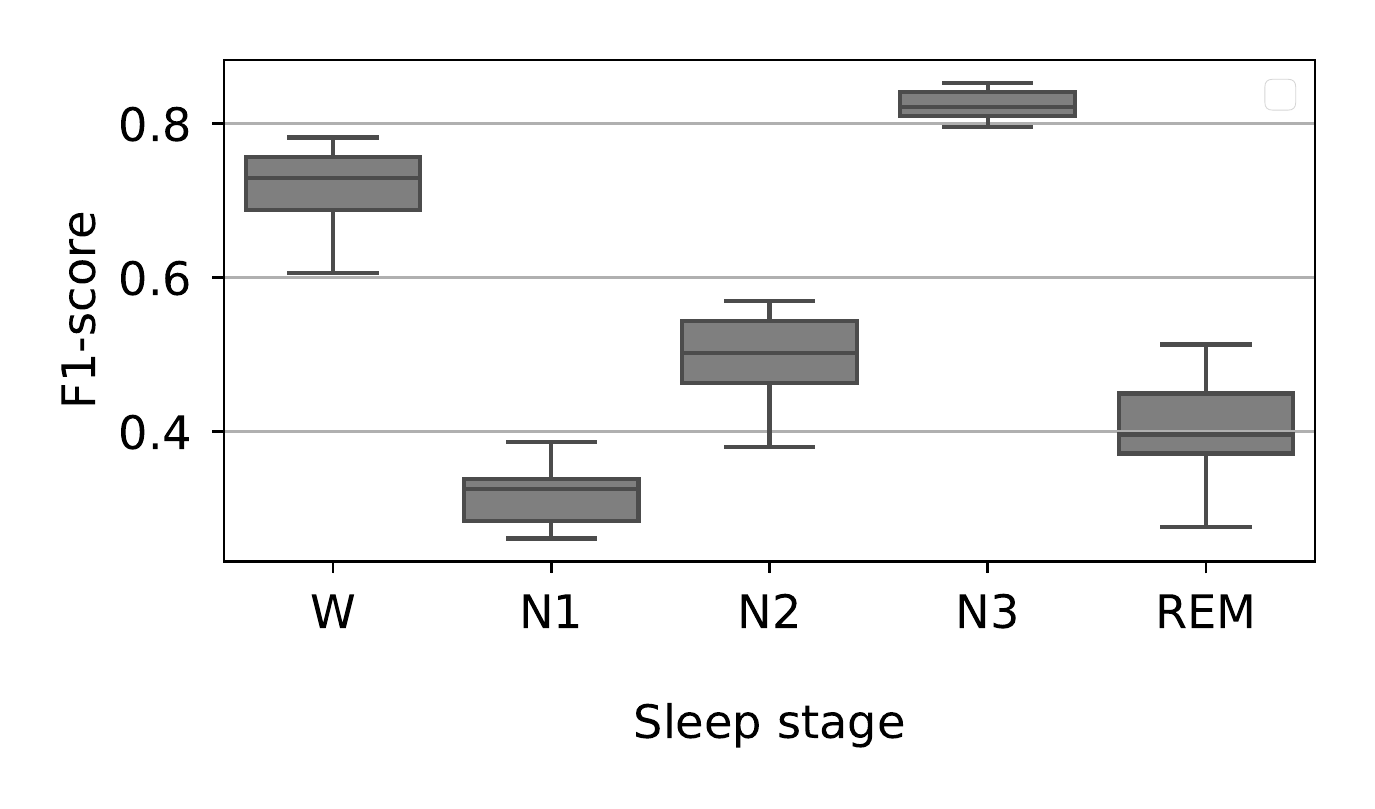}
        \caption{\emph{SleepPhysionet}}
        \label{fig:box_plot_per_class_ref_physionet}
    \end{subfigure}
    \hfill
    \begin{subfigure}[b]{\columnwidth}
        \centering
        \includegraphics[width=\columnwidth]{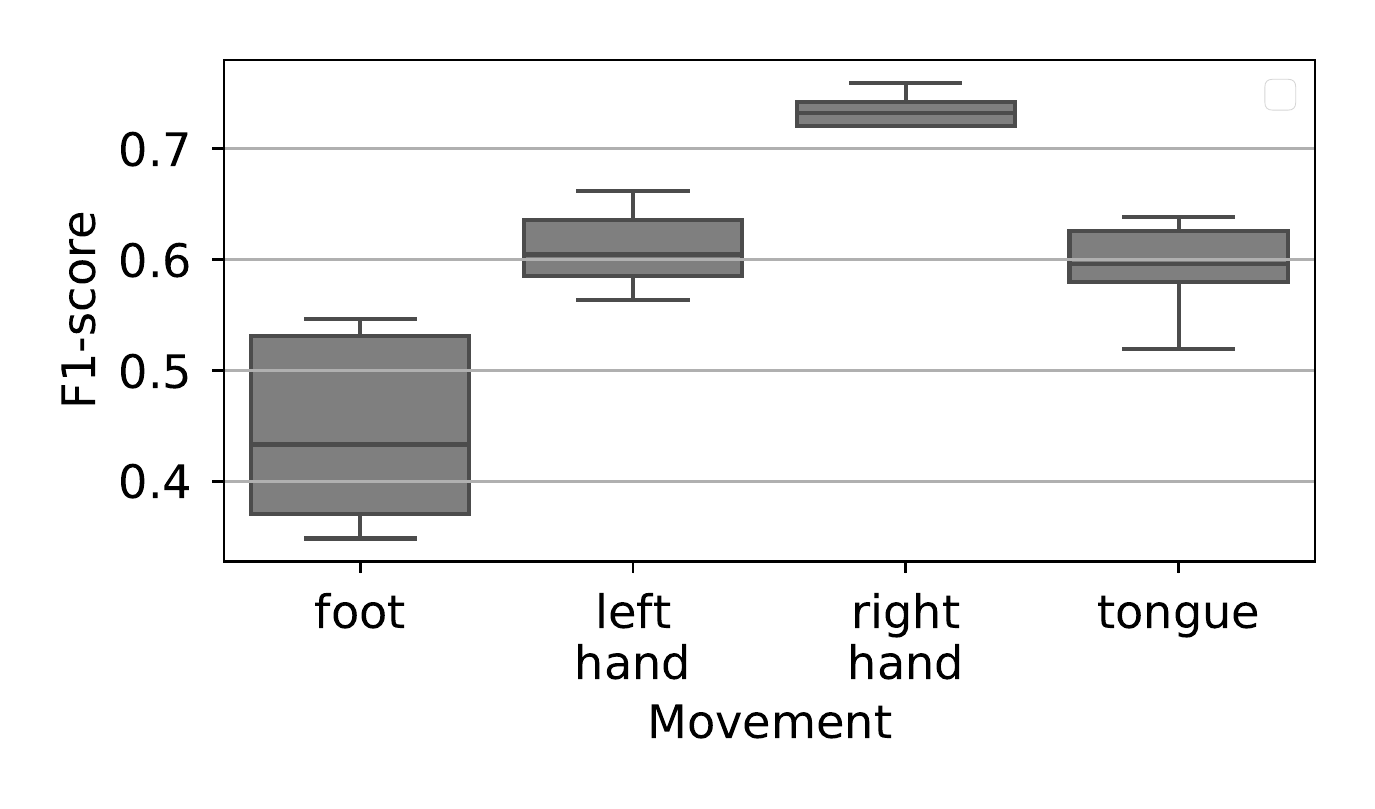}
        \caption{\emph{BCI IV 2a}}
        \label{fig:box_plot_per_class_ref_BCI}
    \end{subfigure}
    \caption{
    Per-class F1-score for a baseline model trained without data augmentation. Boxplots were estimated using $10$-fold cross-validation.}
\end{figure*}

\Cedric{\autoref{fig:freq_boxplot_physionet} shows the F1-score improvements per class when using each frequency domain data augmentation in the sleep staging task.}
First, it can be seen that all frequency data augmentation methods only produce marginal improvements for sleep stages W and N3.
These results can be interpreted in light of the performance of the baseline model presented in \autoref{fig:box_plot_per_class_ref_physionet}. 
Indeed, the highest F1-scores are reached for these classes and it is probably difficult to improve over the representations extracted by the baseline.
\Cedric{On the contrary, REM stage appears as benefiting the most from the frequency domain augmentations.}
\Cedric{This cannot be completely explained by the low baseline accuracy for this stage, since a similar baseline performance is obtained for the N2 stage, which benefits less from data augmentation.}

\paragraph{BCI IV 2a}
\Cedric{Per class results for the BCI task are presented in \autoref{fig:freq_boxplot_BCI}.}
\Cedric{Frequency augmentations seem to bring the smallest improvements for right hand movements, which might be due to the high performance reached by the baseline for this class (\lcf \autoref{fig:box_plot_per_class_ref_BCI}).}
\Cedric{\texttt{FTSurrogate} consistently leads to larger performance improvements than other augmentations for 3 out of 4 classes.}

\subsection{\Cedric{Conclusion of frequency augmentations experiments}}

\Cedric{In both sleep stage classification and motor imagery experiments, the \texttt{FTSurrogate} augmentation seems to consistently lead to the most significant performance boost.}
\Cedric{For instance, this augmentation reached 45\% relative improvement compared to the baseline when training on a small portion of the BCI dataset.}
\Cedric{Although performance boosts are not as impressive in our sleep stage classification experiments, we demonstrate that using \texttt{FTSurrogate} can sometimes be equivalent to training on a dataset four times larger.}
\Cedric{Using the maximum magnitude for this augmentation, as proposed in the original publication \cite{schwabedal_addressing_2019}, seems to yield the best results in both datasets.}
\Cedric{Surprisingly, \texttt{FrequencyShift} seems to be useful not only for sleep stage classification, but also for BCI, with improvements sometimes comparable to \texttt{FTSurrogate}.}
\Cedric{The \texttt{BandstopFiltering} augmentation, however, seems beneficial only for BCI applications and not for sleep staging.}

\section{Spatial domain augmentations}
\label{sec:spatial-tfs}

In this section we study augmentations exploiting the sensors spatial positions.
\subsection{\Cedric{Rationale of spatial augmentations}}

\begin{mdframed}[style=mystyle, frametitle=Channels symmetry]
    \Cedric{The \texttt{ChannelsSymmetry} augmentation, proposed in \cite{deiss_hamlet_2018}, simulates a swap between EEG sensors placed on the right and left hemispheres.}
\end{mdframed}

\Cedric{More practically, this is obtained by permuting the rows of the input signal $X$ in a particular way.}
\Cedric{Each row or channel of $X$ corresponds to a known sensor position, \emph{e.g.} $[C3, C4, F3, F4, O1, O2]$ in a standard 10-20 system.}
\Cedric{In this setting, odd indices correspond to the left hemisphere and even indices to the right hemisphere.}
\Cedric{Hence, we can obtain the augmented signal by swapping rows corresponding to the same electrodes in left and right sides: ${C3 \leftrightarrow C4}$, ${F3 \leftrightarrow F4}$, ${O1 \leftrightarrow O2}$.}

\Cedric{This augmentation is motivated by the observation that} several brain activities
monitored using EEG involve a sagittal plane symmetry (left-right).
For example, it has been evidenced that tongue movements stem from an activation of the primary and supplementary sensorimotor areas without any significant lateralization \cite{watanabe_human_2004}.
%
\Cedric{Yet, motor imagery tasks involving hand movements are strongly lateralized and dominated by contralateral activations~\cite{penfield_epilepsy_1954},}
\Cedric{suggesting that applying the \texttt{ChannelsSymmetry} transformation in such a context would degrade model performance.}

\begin{mdframed}[style=mystyle, frametitle=Channels dropout]
    \Cedric{The \texttt{ChannelsDropout} augmentation, initially proposed in \cite{saeed_learning_2020}, randomly sets some channels of the EEG recording to zero with a given probability $\pdrop$.}
\end{mdframed}

\Cedric{More precisely, for $X \in \mathbb{R}^{C \times T}$ an EEG window of $T$ samples collected on $C$ channels, the augmented signal takes the form
\begin{equation*}
    \mathtt{ChannelsDropout}[X]_c := d_c \cdot X_c,
\end{equation*}
where $d_c$'s are sampled from a Bernoulli distribution $\mathcal{B}$ of probability $\pdrop$, and $X_c$ denotes the $c^{th}$ row of $X$  corresponding to channel $c \in \{1, \dots, C\}$.}

\Cedric{As the widely used dropout layer \cite{srivastava_dropout_2014}, the motivation of this augmentation is to prevent the model from relying too heavily on a given input channel, which could lead to overfitting and poor generalization on different datasets. The transformation naturally improves the robustness of the model to corrupted channels, which is}
a major hurdle for the analysis of EEG signals.
For example, in polysomnography, changes in the subject's position during sleep might result in a loss of contact between several electrodes and the scalp.
Beyond sleep applications, the spread of mobile wearable EEG devices raises new challenges, as they are more prone to noise and missing channels \cite{banville2021robust}.
Finally, the EEG and machine learning communities consider with great interest the question of transferability across datasets, which raises major challenges regarding inconsistent numbers of channels or channels ordering~\cite{rodrigues-etal-RPA:19}.
%
%
%
\Cedric{This augmentation hence drives the model to learn}
from the global information available from all channels instead of relying on a single one.

\begin{mdframed}[style=mystyle, frametitle=Channels shuffle]
    \Cedric{The \texttt{ChannelsShuffle} augmentation, also proposed in \cite{saeed_learning_2020}, consists in randomly permuting the rows of EEG input matrices.}
\end{mdframed}

\Cedric{Given an input signal $X \in \mathbb{R}^{T \times C}$, the augmented signal is defined as}
\begin{equation*}\label{eq:channel-shuffle}
    \mathtt{ChannelsShuffle}[X]_c := X_{\tau(c)},
\end{equation*}
where $\tau$ is uniformly sampled from all the possible permutations
\Cedric{of a random subset of channels $I$.}
Although all channels are shuffled in the original formulation proposed in \cite{saeed_learning_2020} ($I=\{1, \dots, C\}$), we implement this augmentation with a variable subset of permuted channels:
\begin{equation*}
  I=\{c | s_c=1, s_c \sim \mathcal{B}(\pshuffle)\},
\end{equation*}
\Cedric{where $\mathcal{B}$ is a Bernoulli distribution and $\pshuffle$ is the probability of adding each channel to the permutation set $I$.}
\Cedric{The parameter $\pshuffle$ hence sets the magnitude of this transformation as it}
defines how strongly the input signals will be distorted.
\Cedric{The original formulation from \cite{saeed_learning_2020} can be obtained by choosing $\pshuffle=1$.}

In addition to helping the transfer to datasets with different channels ordering, this augmentation
induces invariance to the absolute and relative positioning of EEG sensors, since it prevents the decision function from relying on it.
Such a transformation can make sense for several EEG classification tasks for which the precise localization of the cerebral activity is not strongly predictive, such as sleep staging.
Indeed, sleep experts hardly consider the sensors position of the channel they observe (\eg Fpz-Oz) but rather rely on the waveforms and spectral characteristics (\eg theta activity, K-complex).
If this augmentation seems well suited for the aforementioned task, it is expected to be less efficient in a context where sensors positions and source localization features have a greater impact, such as BCI.

\begin{mdframed}[style=mystyle, frametitle=Sensors rotations]
    \Cedric{The \texttt{SensorsRotation} augmentation, proposed in \cite{krell_rotational_2017}, approximates what would have been recorded with a device slightly rotated by a random angle along a given axis (x, y or z).}
\end{mdframed}

\Cedric{To achieve this, a random angle $\theta$ is drawn for each new input $X$ within a chosen range $[-\trot, \trot]$.}
\Cedric{Then, the sensors 3D coordinates in a standard 10-20 montage are rotated by $\theta$ along the desired axis.}
\Cedric{Finally, the electrical potentials in $X$ are interpolated from the original sensor positions to the new rotated ones.}
\Cedric{While in \cite{krell_rotational_2017} a radial-basis functions interpolator is used, we implemented this transformation using spherical splines from the \textsc{MNE-Python} library~\cite{gramfort_meg_2013}, as commonly done for bad EEG channels preprocessing \cite{perrin_spherical_1989}. Following the MNE software head coordinate convention, the X axis goes from the left to the right ear, the Y axis goes from the back of the head to the nose, while the Z axis goes upwards.}
\Cedric{The magnitude of this transformation is set through the maximum rotation angle $\trot$.}

\Cedric{The main idea of this augmentation is to promote robustness to perturbations of the EEG sensors positions.}
\Cedric{It is motivated by the fact that,}
between different recording sessions, the \Cedric{EEG} cap can move over the subjects head, resulting in slightly shifted sensors locations.
%
%
Compared to \texttt{ChannelsShuffle}, which encourages \emph{global} invariance to electrodes positions, \texttt{SensorsRotation} induces robustness to small and local variations.

\subsection{\Cedric{Empirical comparison of spatial augmentations}}
\label{sec:spatial_results}

\subsubsection{Parameters selection}

The parameters listed in \autoref{table:spatial_param_search} control the magnitude of the transformations, namely: the probability to drop channels $\pdrop$, the probability to shuffle channels $\pshuffle$ and the angle of rotation $\trot$ respectively for \texttt{ChannelsDropout}, \texttt{ChannelsShuffle} and \texttt{SensorsRotations}.
Regarding the sensors rotations, we restricted the range of possible angles $\trot$ to $[0, 30]$ degrees as done in \cite{krell_rotational_2017}.
The magnitude of the \texttt{ChannelsSymmetry} augmentation cannot be adjusted.

\begin{table*}[ht]
    \centering
    \small
    \begin{tabular}{c c c c c c} 
        \hline\\
        Augmentation & Parameter & Interval & Unit & Best value (sleep staging) & Best value (BCI)\\
        \hline
        \texttt{ChannelsDropout} & $\pdrop$ & $[0, 1]$ & - & $0.4$ & $1$ \\
        \texttt{ChannelsShuffle} & $\pshuffle$ & $[0, 1]$ & - & $0.8$ & $0.1$ \\
        \texttt{SensorXRotations} & $\trot$ & $[0, 30]$ & degree & $25^o$ & $3^o$\\
        \texttt{SensorYRotations} & $\trot$ & $[0, 30]$ & degree & $9^o$ & $12^o$\\
        \texttt{SensorZRotations} & $\trot$ & $[0, 30]$ & degree & $30^o$ & $3^o$\\
        \hline
    \end{tabular}
    \normalsize
    \caption{Potential and selected values for the adjustable parameter of each spatial domain augmentation.}
    \label{table:spatial_param_search}
\end{table*}

\begin{figure}[h!]
     \centering
     \begin{subfigure}[b]{\columnwidth}
         \centering
         \includegraphics[width=\columnwidth]{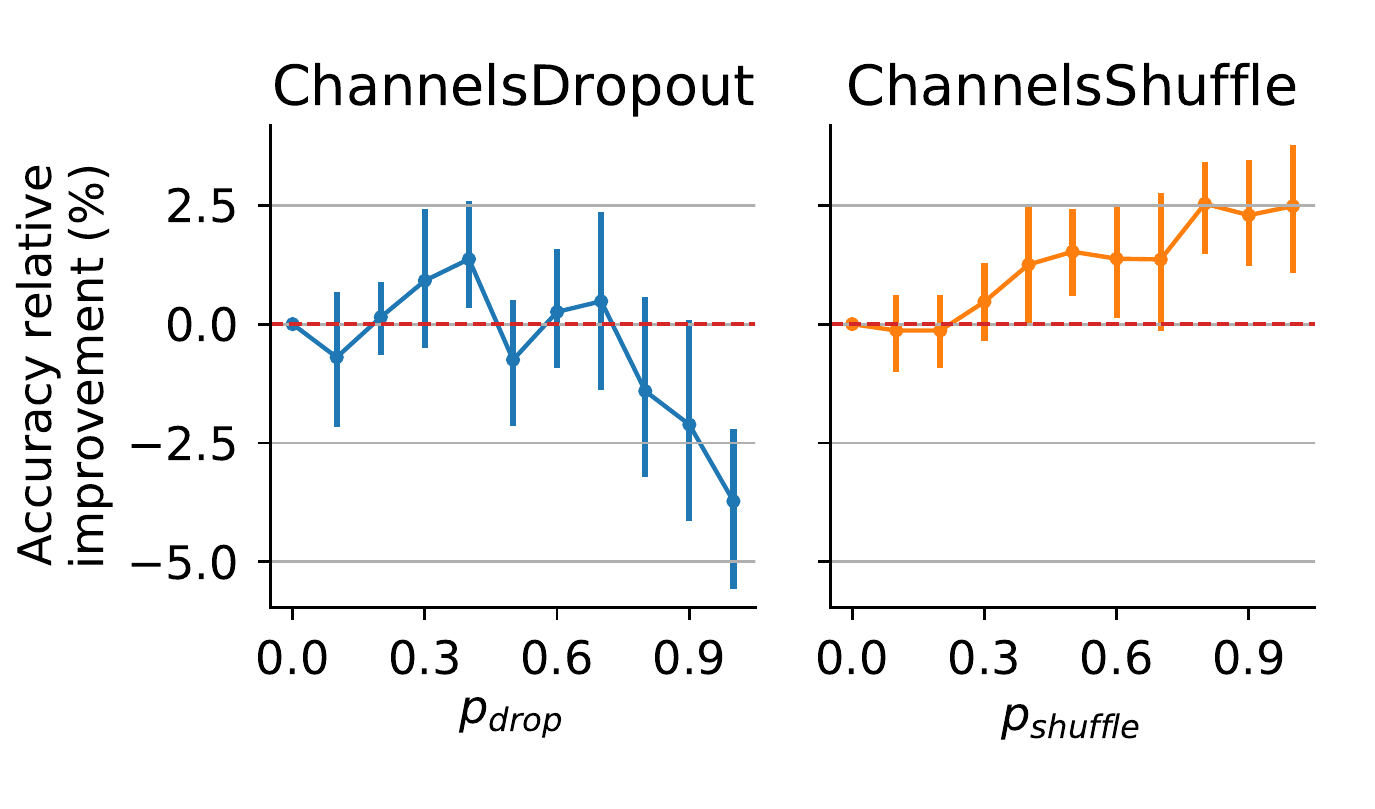}
         \caption{\emph{SleepPhysionet}}
         \label{fig:space_param_search_physionet}
     \end{subfigure}
     \begin{subfigure}[b]{\columnwidth}
         \centering
         \includegraphics[width=\columnwidth]{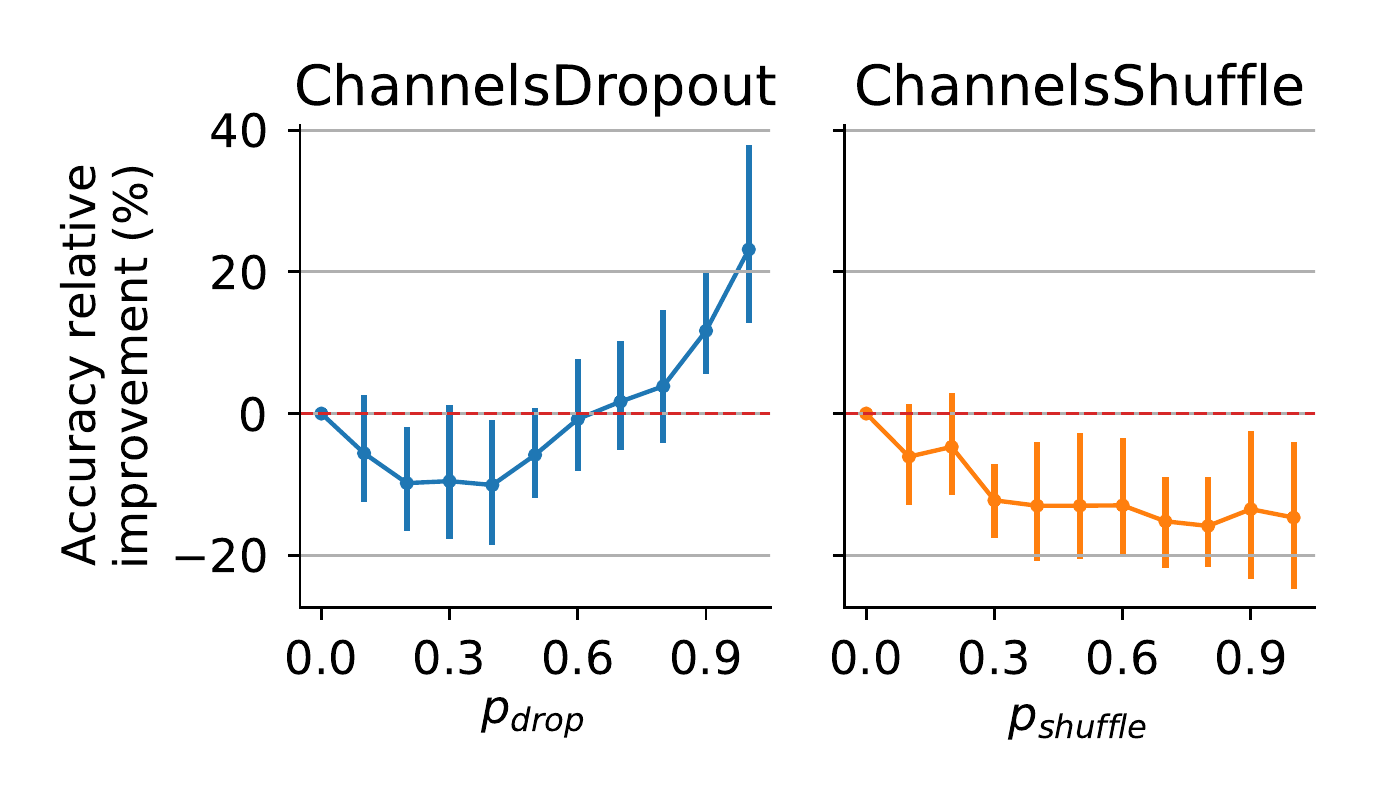}
         \caption{\emph{BCI IV 2a}}
         \label{fig:space_param_search_BCI}
     \end{subfigure}
     \caption{Spatial augmentations parameters selection on the \emph{SleepPhysionet} (a) and \emph{BCI IV 2a} (b) datasets. Models were trained on respectively 350 and 60 windows using augmentations parametrized with 10 different linearly spaced values.
     Validation accuracies are reported relatively to a model trained without data augmentation.
     The error bars correspond to the 95\% confidence intervals based on a 10-fold cross-validation.}
     \label{fig:spatial_param_search}
\end{figure}

\begin{figure*}
    \centering
     \begin{subfigure}[b]{1\textwidth}
         \centering
         \includegraphics[width=0.9\textwidth]{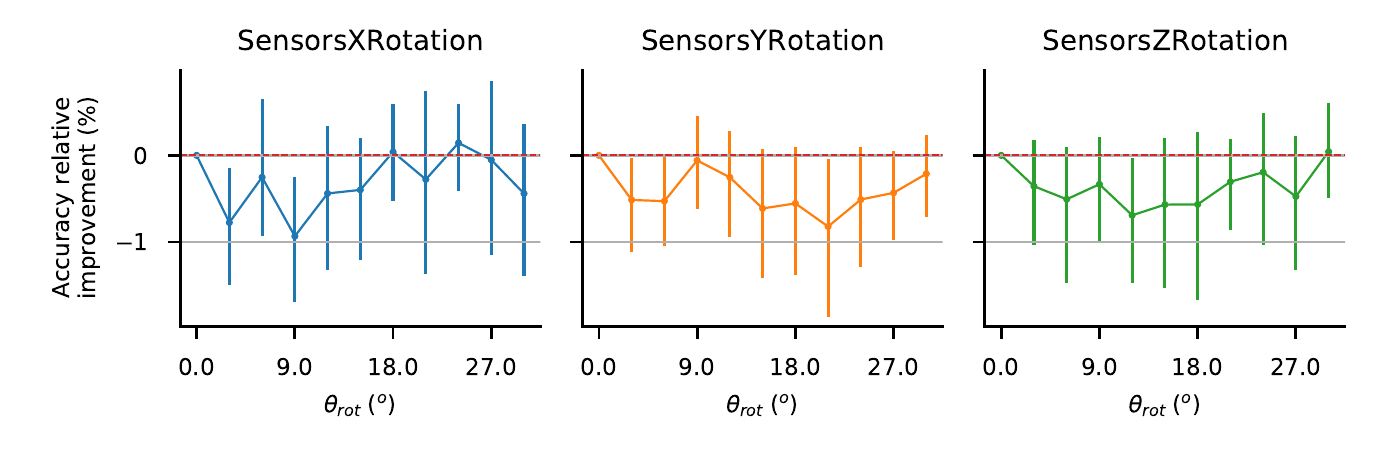}
         \caption{\emph{SleepPhysionet}}
         \label{fig:rot_param_search_physionet}
     \end{subfigure}
    \begin{subfigure}[b]{1\textwidth}
         \centering
         \includegraphics[width=0.9\textwidth]{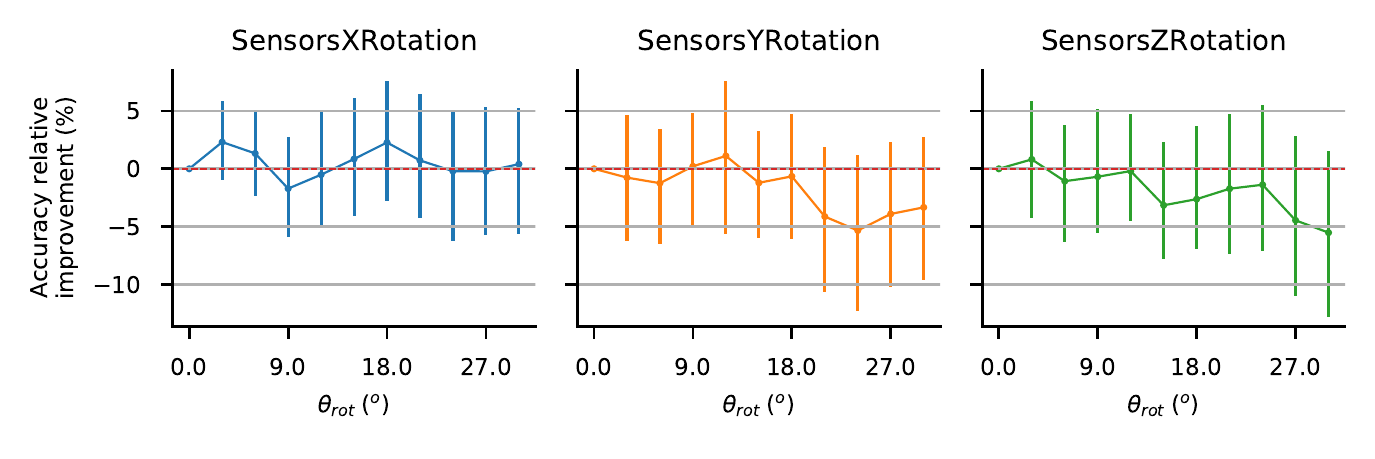}
         \caption{\emph{BCI IV 2a}}
         \label{fig:rot_param_search_BCI}
     \end{subfigure}
    \caption{Rotational augmentations parameters selection on the \emph{SleepPhysionet} (a) and \emph{BCI IV 2a} (b) datasets. Models were trained on respectively 350 and 60 windows using augmentations parametrized with 10 different linearly spaced values.
    Validation accuracies are reported relatively to a model trained without data augmentation.
    The error bars correspond to the 95\% confidence intervals based on a 10-fold cross-validation.}
    \label{fig:rot_param_search}
\end{figure*}

\paragraph{SleepPhysionet}
The results of the grid search for the parameters of spatial augmentations are presented in \Autoref{fig:space_param_search_physionet, fig:rot_param_search_physionet}. 
We can see that \texttt{SensorsRotations} consistently lead to poor performances.
This can be explained by the scarce number of EEG sensors available in this dataset, resulting in imprecise interpolation.
This same reason might also explain why high probabilities of dropping channels in \texttt{ChannelsDropout} appear to be detrimental to learning, the best results being obtained with $\pdrop=0.4$.
Indeed, as there are only two channels in this dataset, a probability of $\pdrop=0.7$ would erase all channels for 1 out of 4 windows on average (the probability to augment $\paug=0.5$, multiplied by $\pdrop$ squared).
On the contrary, for \texttt{ChannelsShuffle}, higher probability to shuffle channels yields the best results.
This was to be expected, since sleep stage information is not very spatially localized.

\paragraph{BCI IV 2a}
As shown in \autoref{fig:space_param_search_BCI}, the magnitude of the spatial domain augmentations impacts the performance on the motor imagery task in a very different way than on the sleep staging task.
Indeed, the patterns for \texttt{ChannelsDropout} and \texttt{ChannelsShuffle} are reversed here: larger values of $\pdrop$ yield better performances, while \texttt{ChannelsShuffle} is consistently harmful, with stronger shuffling probabilities yielding the worst performances.
The order of magnitude of the impact of these augmentations is also different compared to the sleep staging case, with up to 20\% improvement for \texttt{ChannelsDropout}.
Moreover, it is surprising to obtain the best results with a probability $\pdrop=1$, given that it corresponds to all channels being dropped for 1 out of 2 windows.
This might indicate that our model is overfitting the data, since gradients computed from fully dropped examples only update the biases of the model.
Regarding the rotational augmentations, parameter selection results are depicted in \autoref{fig:rot_param_search_BCI}. \Cedric{Due to very large error bars and mean values close to zero, practical usefulness of random channel rotations is hard to assess here.}


\subsubsection{Learning curves}
\label{sec:spatial-learning-curve}

\paragraph{SleepPhysionet}

\begin{figure*}[ht]
     \centering
     \begin{subfigure}[b]{\columnwidth}
         \centering
         \includegraphics[width=\columnwidth]{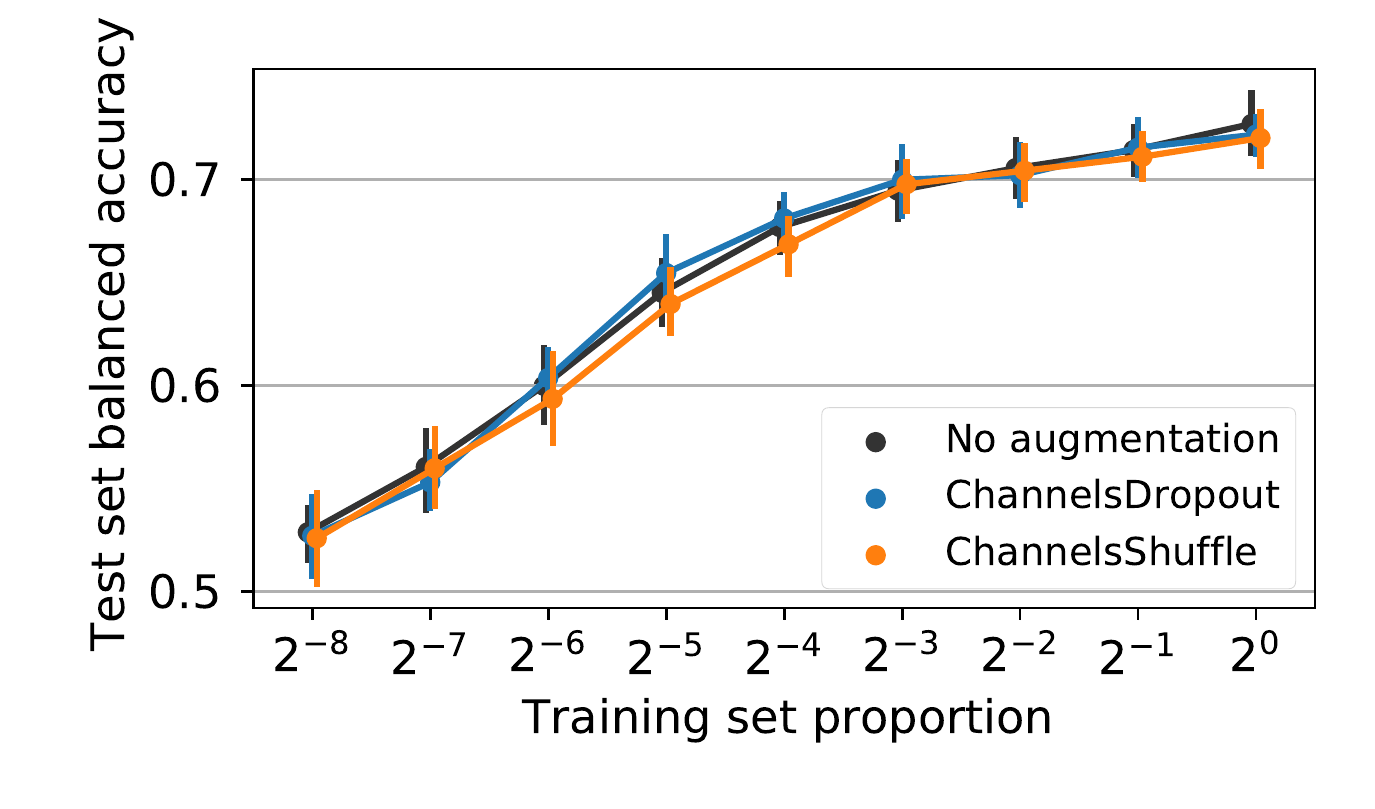}
         \caption{\emph{SleepPhysionet}}
         \label{fig:spatial_LRC_physionet}
     \end{subfigure}
     \hfill
     \begin{subfigure}[b]{\columnwidth}
         \centering
         \includegraphics[width=\columnwidth]{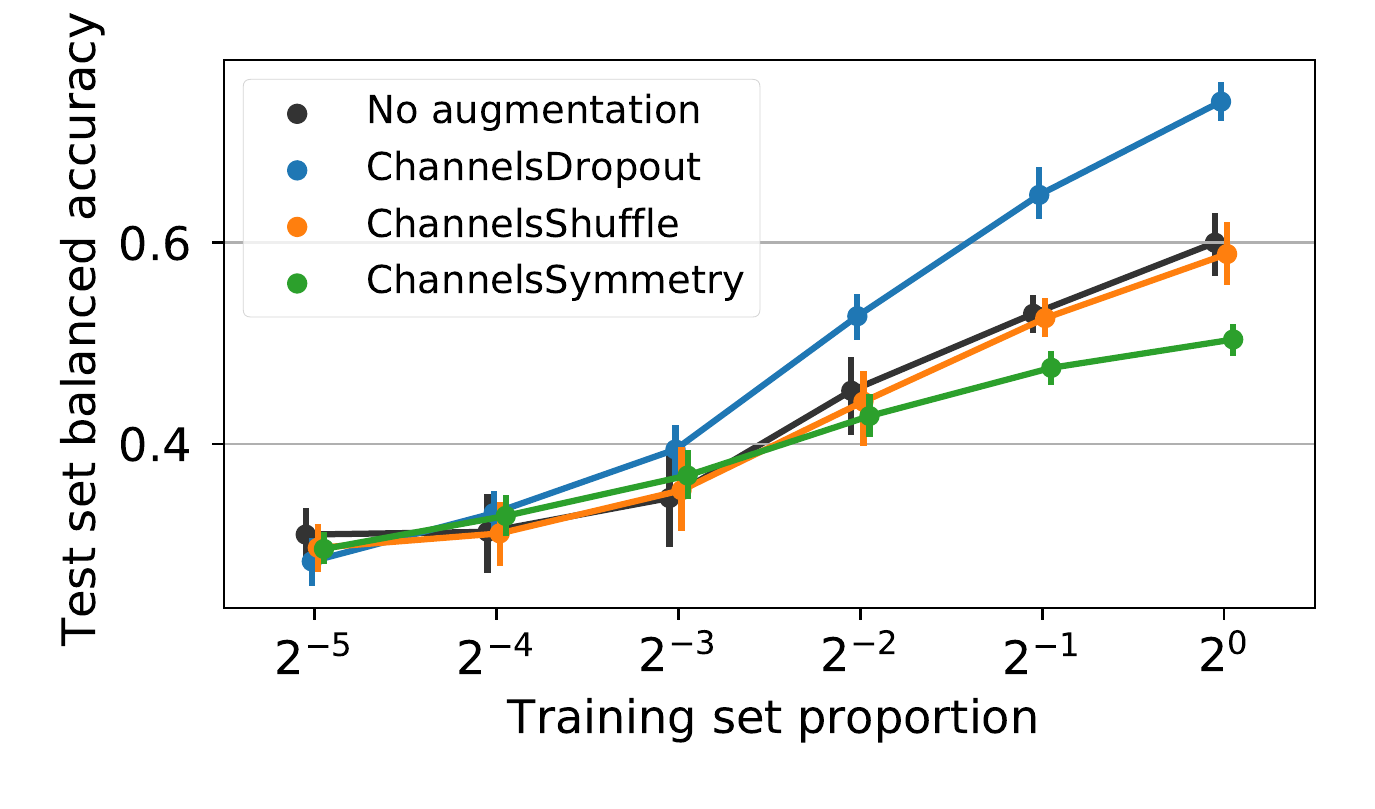}
         \caption{\emph{BCI IV 2a}}
         \label{fig:spatial_LRC_BCI}
     \end{subfigure}
     \caption{Learning curves for spatial domain augmentations (except rotations) along with the baseline trained with no augmentation. For each transformation, the same model is trained on fractions of the dataset of increasing size. After each training, the average balanced accuracy score on the test set is reported with error bars representing the 95\% confidence intervals estimated from 10-fold cross-validation.}
     \label{fig:spatial_LRC}
\end{figure*}

The learning curves of spatial domain augmentations in sleep staging are plotted on \autoref{fig:spatial_LRC_physionet}.
All augmentations seem to globally perform on par with the baseline,
\Cedric{suggesting they}
are not particularly relevant for sleep stage classification.
\Cedric{These poor results could also be explained by the fact that the \emph{SleepPhysionet} dataset comprises only 2 EEG electrodes in the sagital plane.}
Note that,
\Cedric{for this reason,}
the \texttt{ChannelsSymmetry} augmentation was omitted for these experiments as it would correspond to the Identity mapping here.

\paragraph{BCI IV 2a}

\begin{figure}[h!]
    \centering
    \includegraphics[width=\columnwidth]{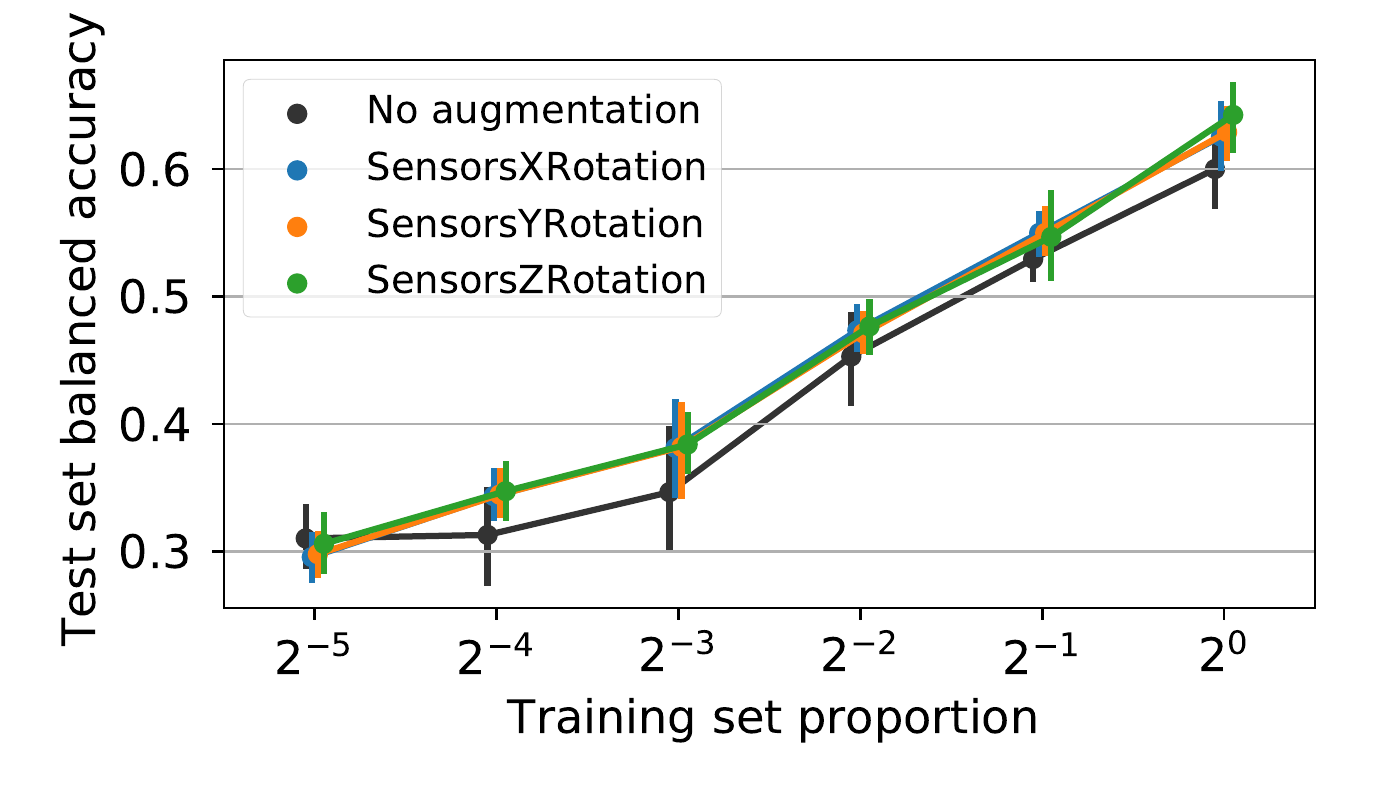}
    \caption{Learning curves for sensors rotations augmentations on the \emph{BCI IV 2a} dataset, along with the baseline trained with no augmentation. For each transformation, the same model is trained on fractions of the dataset of increasing size. After each training, the average balanced accuracy score on the test set is reported with error bars representing the 95\% confidence intervals estimated from 10-fold cross-validation.}
    \label{fig:rot_LRC_BCI}
\end{figure}

The learning curves obtained with spatial data augmentations are presented on \autoref{fig:spatial_LRC_BCI}.
They confirm that mixing channels up with \texttt{ChannelsShuffle} and \texttt{ChannelsSymmetry}
might be detrimental for motor imagery tasks.
\Cedric{These results confirm that the spatial information in multivariate EEG recordings is crucial for BCI.}
On the contrary, \texttt{ChannelsDropout} significantly enhance the performance, specially
\Cedric{on larger training sets.}
Unlike the two previous spatial augmentations, \texttt{ChannelsDropout} helps to learn more robust motor imagery features by hiding part of the spatial information without misleading the model.
\Cedric{Finally, its seems \texttt{SensorsRotation} augmentations can be helpful although performance improvements depicted in \autoref{fig:rot_LRC_BCI} are not statistically significant, as already observed in \autoref{fig:rot_param_search_BCI}.}


\subsubsection{Per class analysis}
\label{subsubsec:spatial_perclass}

\paragraph{SleepPhysionet}

\begin{figure*}[ht]
     \centering
     \begin{subfigure}[b]{\columnwidth}
         \centering
         \includegraphics[width=\columnwidth]{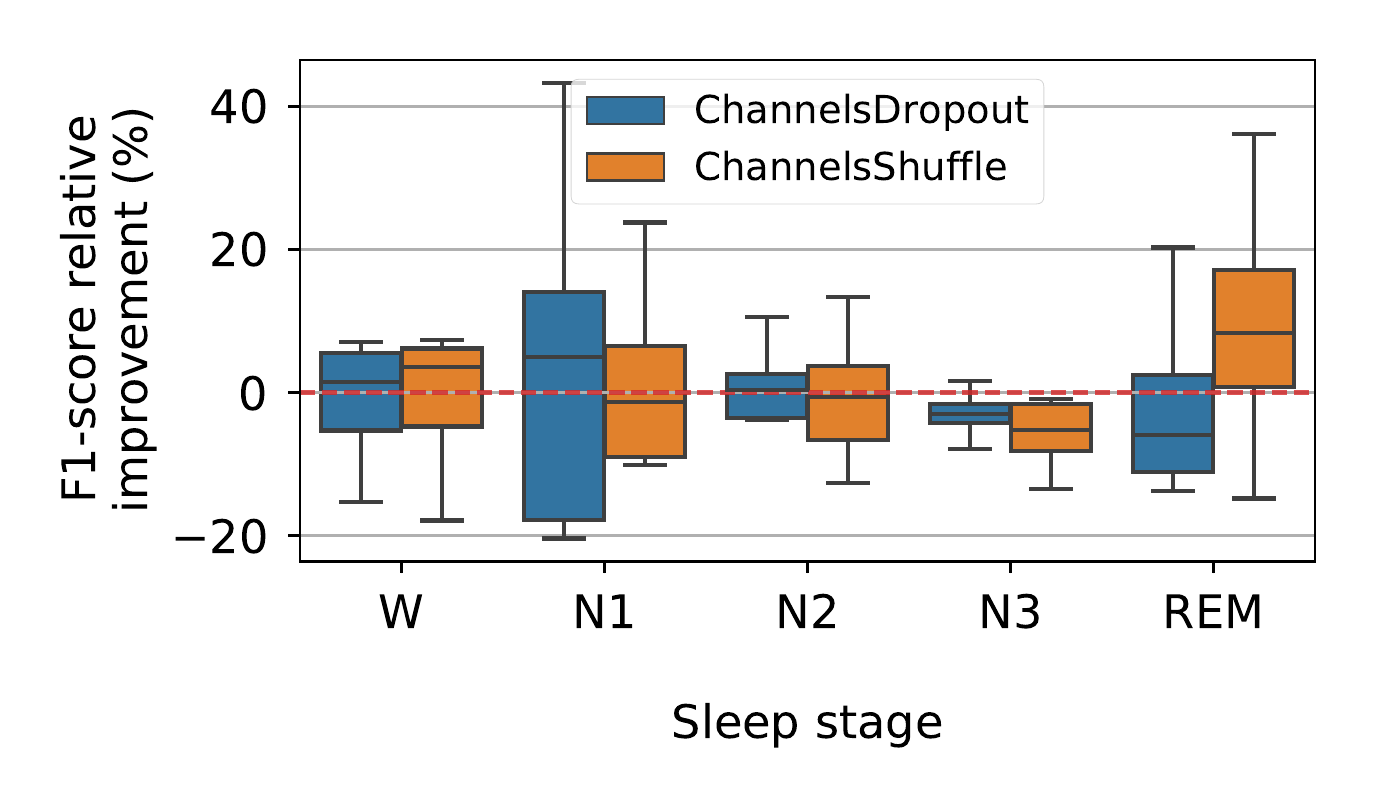}
         \caption{\emph{SleepPhysionet}}
         \label{fig:spatial_boxplot_physionet}
     \end{subfigure}
     \hfill
     \begin{subfigure}[b]{\columnwidth}
         \centering
         \includegraphics[width=\columnwidth]{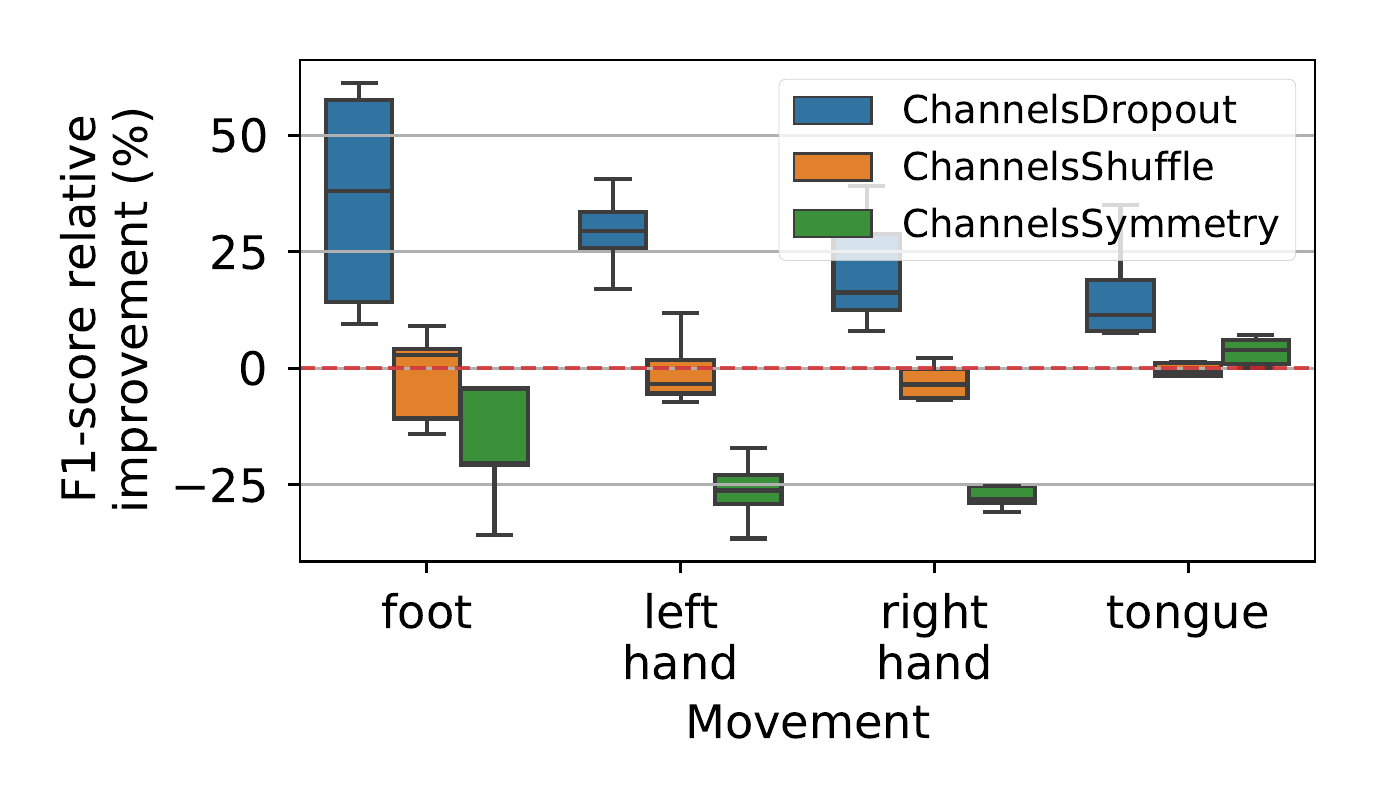}
         \caption{\emph{BCI IV 2a}}
         \label{fig:spatial_boxplot_BCI}
     \end{subfigure}
     \caption{Per-class F1-score for spatial domain transformations (except for rotations). Scores are reported as relative improvement over a baseline trained without data augmentation. Models were trained on $180$ and $230$ time windows for \emph{SleepPhysionet} and \emph{BCI IV 2a} datasets respectively. Boxplots were estimated using $10$-fold cross-validation.}
\end{figure*}

While \autoref{fig:spatial_boxplot_physionet} confirms that spatial augmentations have no significant impact on the classification of stages W, N1 and N2, it also brings more nuance to the previous results from \autoref{fig:spatial_LRC_physionet}.
Indeed, it seems that
dropping channels can significantly harm the recognition of N3 and REM stages,
\Cedric{suggesting they}
are more easily identified on one of the two available channels.
Likewise, shuffling seems to degrade the performance for the N3 stage, while yielding 10\% median improvement for REM stages.
This might indicate that N3 stages are partly characterized by spatial patterns which are lost when channels are shuffled.
It also seems to confirm that REM stages are not localized and rather correspond to a global brain activity, similar to the awake state.

\begin{figure}
    \centering
    \includegraphics[width=\columnwidth]{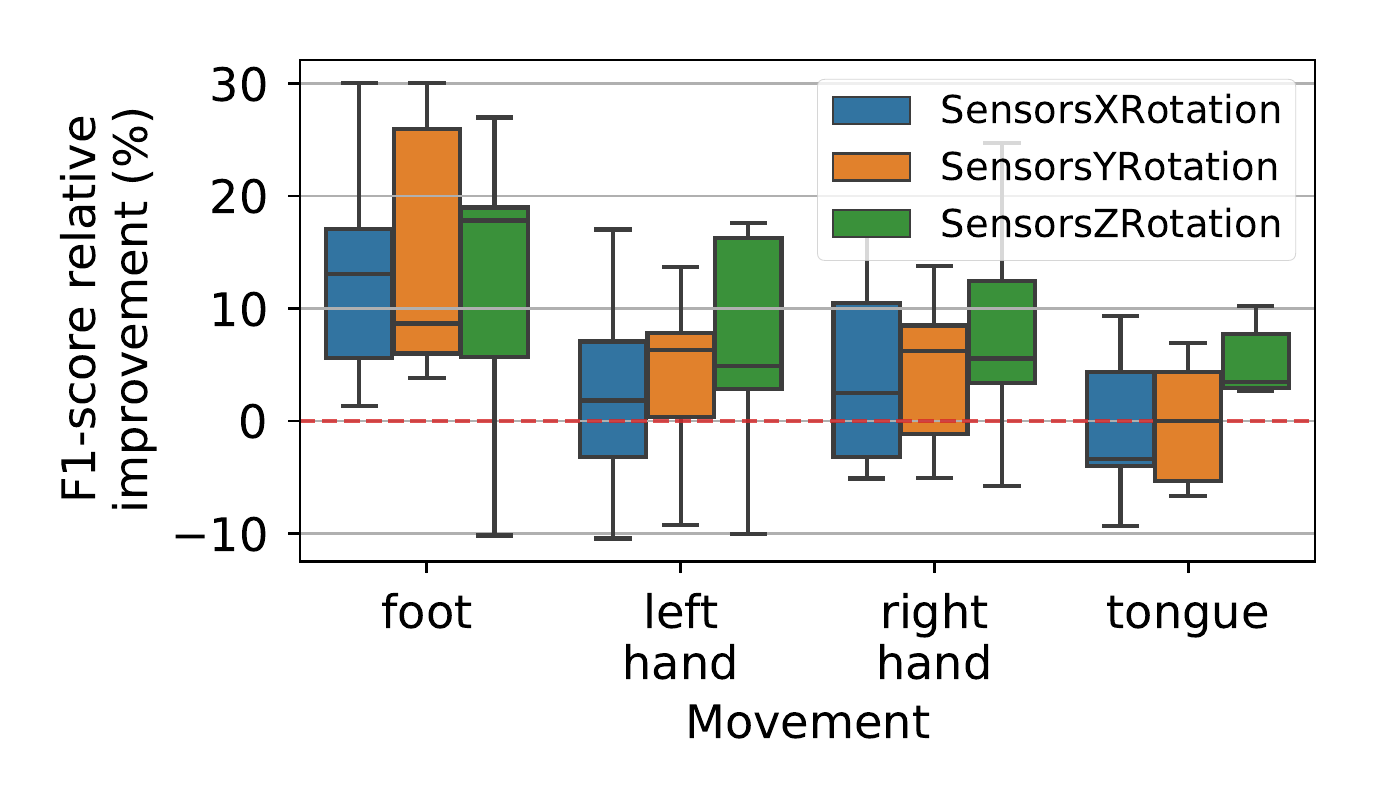}
    \caption{Per-class F1-score for rotation transformations. Scores are reported as relative improvement over a baseline trained without data augmentation. Models were trained on $230$ time windows of the \emph{BCI IV 2a} dataset. Boxplots were estimated using $10$-fold cross-validation.}
    \label{fig:rot_boxplot_BCI}
\end{figure}

\paragraph{BCI IV 2a}

The results per class for the motor imagery task presented in \autoref{fig:spatial_boxplot_BCI} confirm that \texttt{ChannelsSymmetry} is particularly detrimental \Cedric{to the classification of} right and left hand movements, which are characterised by a heavily lateralised brain activity.
Meanwhile, this augmentation slightly helps to learn to recognize tongue movements, which are associated with non-lateralised brain activities.
In fact, this last class seems to be the least affected by all three augmentations.
\Cedric{These results confirm that tongue movements are not spatially characterized and lateralised as heavily as the other movements considered \cite{watanabe_human_2004}.}
Concerning spatial rotations, \autoref{fig:rot_boxplot_BCI} also helps to nuance the previous aggregated results from \autoref{fig:rot_LRC_BCI}.
Indeed, although boxes' whiskers reach negative values, rotations around the Z axis consistently improve the performance in at least 75\% of cases by a significant amount.

\subsection{\Cedric{Conclusion of spatial augmentations experiments}}

\Cedric{While some spatial domain augmentations lead to promising results on our motor imagery BCI experiments, none of them seems very helpful for sleep stage classification.}
\Cedric{\texttt{ChannelsDropout} appears to be particularly interesting for BCI, leading to up to 25\% accuracy boosts.}
\Cedric{\texttt{SensorsRotations} can also lead to interesting performance increases, specially around axis Z (longitudinal axis pointing up).}
\Cedric{On the contrary, the sensors permutations done in \texttt{ChannelsShuffle} and \texttt{ChannelsSymmetry} can actually harm the performance when training to classify lateralised brain activities.}

\section{\Cedric{General discussion and findings summary}} \label{sec:gen-discussion}

\begin{figure*}[t]
    \begin{subfigure}[b]{\textwidth}
         \centering
         \caption{\emph{SleepPhysionet}}
         \label{fig:gen-comparison-sleep}
         \includegraphics[width=\textwidth]{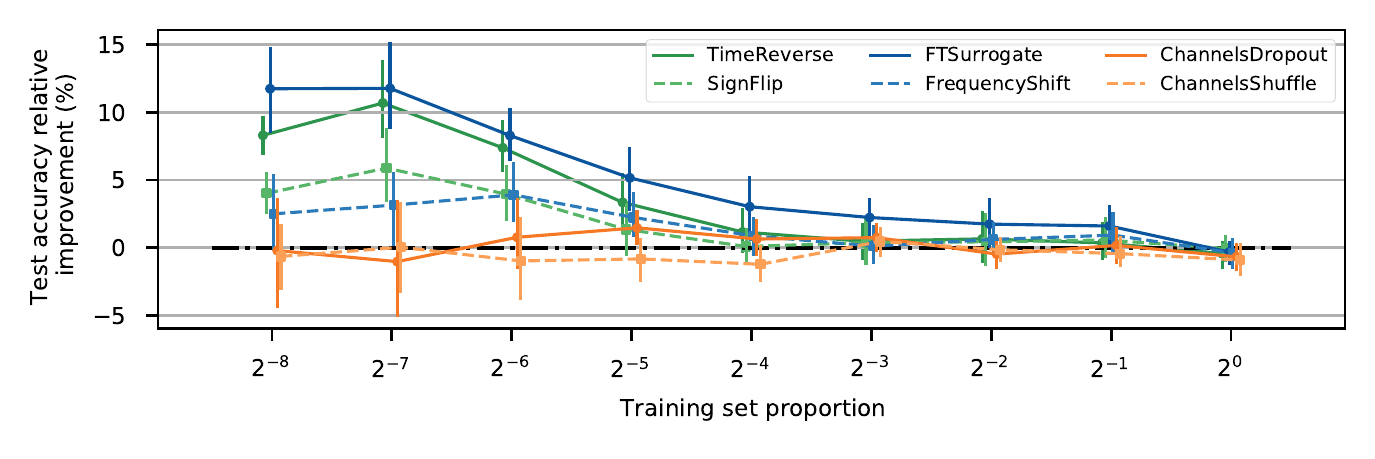}
    \end{subfigure}
    \begin{subfigure}[b]{\textwidth}
         \centering
         \caption{\emph{BCI IV 2a}}
         \label{fig:gen-comparison-bci}
         \includegraphics[width=\textwidth]{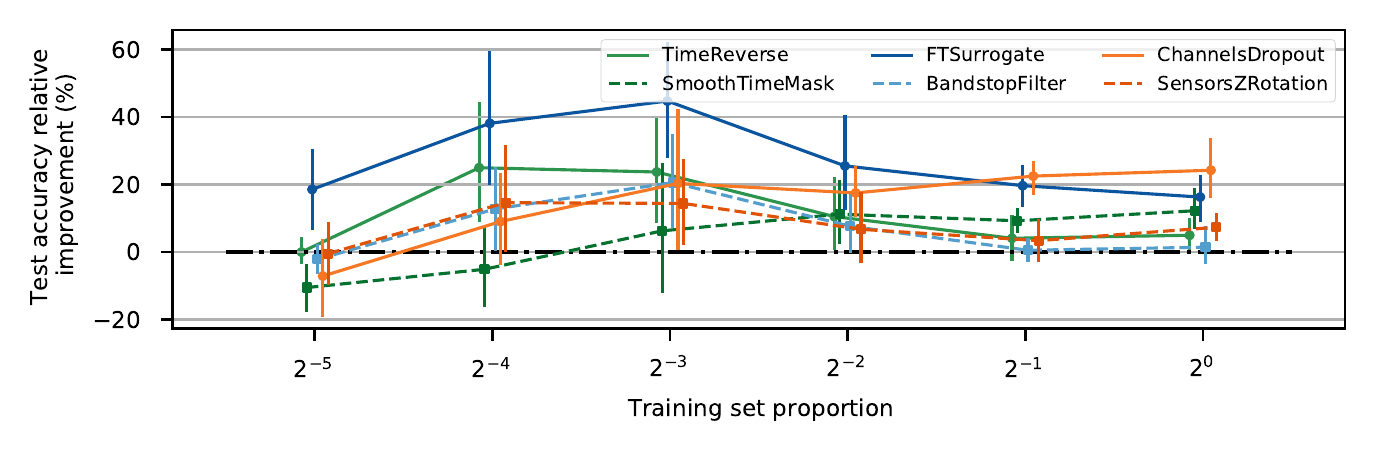}
    \end{subfigure}
    \caption{Comparison between the three groups of augmentations. Each curve corresponds to the same model trained with a different augmentation on fractions of the dataset of increasing size. After each training, the average balanced accuracy score on the test set is computed and reported as an improvement relative to the baseline model trained without data augmentation. Error bars represent the 95\% confidence intervals estimated from 10-fold cross-validation. \textcolor{ForestGreen}{Time} augmentations are plotted in \textcolor{ForestGreen}{green}, \textcolor{MidnightBlue}{frequency} augmentations in \textcolor{MidnightBlue}{blue} and \textcolor{BurntOrange}{spatial} augmentations in \textcolor{BurntOrange}{orange}.
    Full lines with round markers correspond to the best augmentation of each group and dataset, while dashed lines with square markers represent the second best augmentations.}
    \label{fig:gen-comparison}
\end{figure*}

\Cedric{The findings of our experiments are well-summarized on \autoref{fig:gen-comparison}, which shows the learning curves of the two best augmentations of each group for the two datasets studied.}
\Cedric{It becomes apparent from \autoref{fig:gen-comparison-sleep} that time and frequency augmentations are preferable for sleep stage classification tasks.}
\Cedric{In this case, data augmentations mainly help when training on small datasets, leading to improvements between 5-12\%, as opposed to 1-2\% boosts when training on larger training set sizes.}
\Cedric{We did not find that spatial domain augmentations are relevant for sleep stage classification, although this might be due to the low spatial resolution of the two-electrode dataset considered.}
\Cedric{In contrast, \autoref{fig:gen-comparison-bci} indicates that motor imagery can benefit from all three groups of augmentations.}
\Cedric{On smaller training sets, we find the same winning time and frequency augmentations as for sleep staging (\texttt{TimeReverse} and \texttt{FTSurrogate}), while \emph{well chosen} spatial augmentations (\texttt{ChannelsDropout}) lead to the best results on larger training sets.}
\Cedric{In general, the BCI task seems to benefit more from data augmentation than the sleep stage classification task, with performance boosts reaching a 45\% increase in small data regimes.}

\CedricTwo{Importantly, one would like to stress that these gains in predictive performance are obtained with a negligible extra computation time. Indeed data augmentation is carried at the data loading level, which is done asynchronously and in parallel during neural networks' training.}

\section{Conclusion} \label{sec:conclusion}

By allowing to increase the training data during learning, data augmentation limits the need for large annotated datasets that are required to fully leverage the potential of deep learning models.
In this paper we carried out a unified and quasi-exhaustive analysis of existing data augmentation methods for EEG signals.
To this end, we have presented the rationale and assumptions behind each augmentation considered, both from the perspective of the underlying neurophysiology and of the experimental setups.
Overall, our experimental results demonstrate that the use of data augmentation is beneficial for the training of EEG classifiers, 
both for sleep staging and BCI tasks.
%

Our experiments
\Cedric{on two very different datasets}
demonstrate the importance of the selection of both the right transformation and magnitude for each different type of task considered.
While time-frequency transformations appear to be preferable for sleep stage classification, spatial augmentations also seem competitive for motor imagery tasks.
Moreover, our per-class analysis allowed to identify structural differences between different imagined actions and sleep stages, thus also demonstrating the descriptive usefulness of augmentations.
These results also align with the claims and experiments presented in \cite{rommel2021cadda}, showing the relevance of class-dependant data augmentation for neuroscience predictive tasks.

While this study is not completely exhaustive and could be enriched by the addition of other datasets and new augmentations, we believe that the methodological framework presented, along with our reproducible code\footnote{\url{https://github.com/eeg-augmentation-benchmark/eeg-augmentation-benchmark-2022}}, should allow to conduct similar analysis on any other EEG dataset and augmentation.
We believe that this systematic analysis of EEG data augmentation will help practitioners improve their predictive models and foster new research works aiming to better understand augmentation methods for EEG signals.

\ack

We would like to thank Apolline Mellot and Hubert J. Banville for their valuable feedback on this manuscript, as well as Martin Wimpff and Bruno Aristimunha for reviewing our augmentation implementations in the braindecode software.
This work was supported by the ANR BrAIN (ANR-20-CHIA-0016) and ANR AI-Cog grants (ANR-20-IADJ-0002).
It was also granted access to the HPC resources of IDRIS under the allocation 2021-AD011012284 and 2021-AD011011172R2 made by GENCI.

\newcommand{\newblock}{}
\bibliographystyle{plainnat}
\bibliography{main}

\newpage
\appendix

\end{document}